\pdfoutput=1

\PassOptionsToPackage{dvipsnames,table}{xcolor}
\documentclass[11pt]{article}

\usepackage[final]{acl}

\usepackage{times}
\usepackage{latexsym}

\usepackage[T1]{fontenc}

\usepackage[utf8]{inputenc}

\usepackage{microtype}

\usepackage{inconsolata}

\usepackage{graphicx}
\usepackage{xcolor}
\usepackage{cite}
\usepackage{url}            
\usepackage{booktabs}       
\usepackage{amsfonts}       
\usepackage{nicefrac}       
\usepackage{microtype}      
\usepackage{blindtext}      
\usepackage{amsmath} 
\usepackage{algorithm}
\usepackage{algpseudocode}
\usepackage{wrapfig}
\usepackage{enumitem}
\usepackage{lscape}
\usepackage{longtable}
\usepackage{adjustbox}
\usepackage{listings}
\newcommand{\lmmseval}{\textsc{LMMs-Eval}}
\newcommand{\lmmsevallite}{\textsc{LMMs-Eval Lite}}
\newcommand{\livebench}{\textsc{LiveBench}}

\definecolor{ntured}{HTML}{D71440}
\definecolor{ntublue}{HTML}{181C62}
\usepackage{marvosym}
\usepackage{makecell}

\usepackage{multirow}
\usepackage{subcaption}
\usepackage{CJKutf8}
\usepackage{xspace} 

\lstdefinestyle{instructformat}{
  basicstyle=\ttfamily\small,
  frame=single,
  breaklines=true,
  breakindent=0pt, 
  keywordstyle=\color{blue},
  keywords={User, GPT},
  xleftmargin=0pt, 
}

\usepackage[capitalise]{cleveref}
\usepackage{verbatim}
\usepackage{bbding}

\usepackage{alltt}
\usepackage{tcolorbox}
\tcbuselibrary{most}
\tcbset{
  aibox/.style={
    width=\textwidth,
    top=10pt,
    colback=white,
    colframe=black,
    colbacktitle=black,
    enhanced,
    center,
    attach boxed title to top left={yshift=-0.1in,xshift=0.15in},
    boxed title style={boxrule=0pt,colframe=white,},
  }
}
\newtcolorbox{AIbox}[2][]{aibox,title=#2,#1}

\title{LMMs-Eval: Reality Check on the Evaluation of Large Multimodal Models}

\author{
  \quad Kaichen Zhang$^{*,1,2}$ 
  \quad Bo Li$^{*,1,2}$ 
  \quad Peiyuan Zhang$^{*,1,2}$ 
  \quad \textbf{Fanyi Pu}$^{*,1,2}$ \\
  \quad \textbf{Joshua Adrian Cahyono} $^{1,2}$
  \quad \textbf{Kairui Hu} $^{1,2}$
  \quad \textbf{Shuai Liu} $^{1,2}$
  \quad \textbf{Yuanhan Zhang} $^{1,2}$ \\
  \quad \textbf{Jingkang Yang} $^{1,2}$
  \quad\textbf{Chunyuan Li}$^{1}$
  \quad\textbf{Ziwei Liu}$^{1,2,}$\textsuperscript{\Letter}\\
  \textsuperscript{1}LMMs-Lab Team
  \quad\textsuperscript{2}S-Lab, NTU, Singapore  \\
{\tt\small\{zhan0564, libo0013, peiyuan.zhang, fpu001, ziwei.liu\}@ntu.edu.sg}\\
}

\begin{document}

\twocolumn[{%
   \renewcommand\twocolumn[1][]{#1}%
   \maketitle
   \vspace{-40pt}
   \begin{center}
    \centering
    \includegraphics[width=1\textwidth]{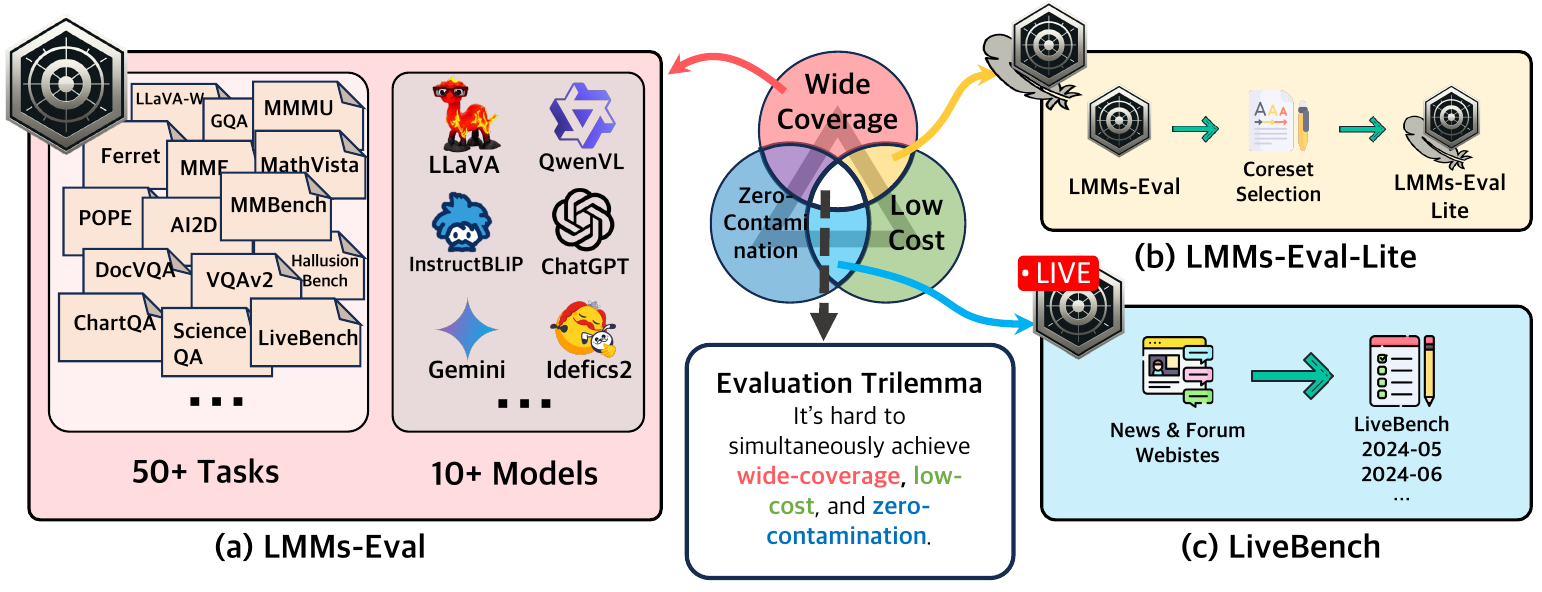}
    \vspace{-10pt}
    \captionof{figure}{To best navigate the trilemma in LMM evaluation benchmarking, we contribute \textbf{(1)} \lmmseval: a unified and standardized multimodal benchmark suite that encompasses over 50 tasks and more than 10 models, ensuring wide coverage; \textbf{(2)} \lmmsevallite: an efficient benchmark set with reliable and aligned results with the time-consuming full-set evaluation, addressing low-cost concerns; \textbf{(3)} \livebench: the evaluation benchmark with the latest information from news and forum websites, aiming to evaluate model's zero-shot generalization ability on most recent events, thereby preventing contamination during evaluations.}
    \label{fig:lmms_eval_banner}
   \end{center}%
  }]
\renewcommand{\thefootnote}{\fnsymbol{footnote}}
\footnotetext[1]{Equal contribution. \\ \textsuperscript{\Letter}Corresponding author.}
\renewcommand*{\thefootnote}{\arabic{footnote}}

\begin{abstract}


    The advances of large foundation models necessitate wide-coverage, low-cost, and zero-contamination benchmarks. Despite continuous exploration of language model evaluations, comprehensive studies on the evaluation of Large Multi-modal Models (LMMs) remain limited. 
    In this work, we introduce \lmmseval, a unified and standardized multimodal benchmark framework with over 50 tasks and more than 10 models to promote transparent and reproducible evaluations.
    Although \lmmseval~offers comprehensive coverage, we find it still falls short in achieving low cost and zero contamination.
    To approach this evaluation trilemma, we further introduce \lmmsevallite, a pruned evaluation toolkit that emphasizes both coverage and efficiency. Additionally, we present Multimodal \livebench~that utilizes continuously updating news and online forums to assess models' generalization abilities in the wild, featuring a low-cost and zero-contamination evaluation approach.
    In summary, our work highlights the importance of considering the evaluation trilemma and provides practical solutions to navigate the trade-offs in evaluating large multi-modal models, paving the way for more effective and reliable benchmarking of LMMs. We opensource our codebase and maintain leaderboard of \livebench at \href{https://github.com/EvolvingLMMs-Lab/lmms-eval}{Github} and \href{https://huggingface.co/spaces/lmms-lab/LiveBench}{LiveBench}. 
\end{abstract}
\section{Introduction}

\paragraph{Good benchmarks guide AI development.} Current large foundational models such as GPT-4~\citep{openai2024gpt4}, Gemini~\citep{geminiteam2024gemini}, Claude~\citep{claude3family2024}, and many others~\citep{qwenvl2024,ormazabal2024reka,mixtral2024,commandrplus2024} have demonstrated transformative capabilities, approaching or surpassing human-level performances in many tasks. In this context, benchmarks become both challenging and crucial to differentiate among the models and detect their weaknesses. 

In the field of language models, exemplary works such as~\citep{liang2022holistic,srivastava2022beyond,eval-harness} aimed to comprehensively assess models across a wide range of dimensions. As generative AI evolves from language-centric to multimodal, a unified evaluation framework and a closer look at existing benchmarks are needed.


\paragraph{Transparent, standardized, and reproducible evaluations are crucial.} We identify that there is so far no unified evaluation protocol in the field of LMM. Model publishers~\citep{liu2023visual, qwenvl2024, dai2023instructblip, zhang2023internlmxcomposer, li2023otter} come up with custom evaluation pipelines, which often differ significantly in data preparation, output postprocessing, and metrics calculation, hindering transparency and reproducibility. To this end, we build a standardized and reliable benchmark suite to assess multimodal models in their entirety with \lmmseval. \lmmseval~covers over 50 tasks in various scenarios to thoroughly assess more than 10 multimodal models with around 30 variants. It offers a standardized evaluation pipeline to ensure transparency and reproducibility. It also comes with a unified interface to facilitate the integration of new models and datasets.


\paragraph{Wide-coverage, low-cost, and zero-contamination benchmark is hard to achieve simultaneously.} We believe it is an impossible triangle to evaluate models with wide coverage and low cost without making the benchmarks susceptible to contamination, as shown in ~\Cref{fig:lmms_eval_banner}. For instance, the Hugging Face OpenLLM leaderboard~\citep{openllmleaderboard} provides an economical way to evaluate language models across a wide range of tasks, but it is also prone to overfitting and contamination. The LMSys Chatbot Arena~\citep{chiang2024chatbot} and AI2 WildVision~\citep{yujie2024wildvisionarena} offer robust and non-contaminated evaluation through real user interactions. However, it is expensive to gather tens of thousands of human preferences.  In this work, we do not break this impossible triangle. Instead, we complement the evaluation landscape of LMMs by introducing \lmmsevallite~ and \livebench~. By covering diverse sets of tasks and pruning unnecessary data instances, \lmmsevallite~ features a low-cost and wide-coverage LMM evaluation. On the other hand, LiveBench gathers the latest information from news and online forums to construct the test data, targeting an economical and generalizable way to do benchmarks.

In summary, we aim to offer a comprehensive view of the evaluations on multimodal models while presenting our observations and solutions. Our paper makes the following contributions:

\textbf{(1)} \lmmseval: a unified multimodal models evaluation suite that covers over 50 tasks and more than 10 models with around 30 sub-variants. With \lmmseval, we aim to streamline and standardize the evaluation process of multimodal models to ensure standardized comparisons between models.

\textbf{(2)} \lmmsevallite: an efficient evaluation set that provides reliable and aligned results with the time-consuming full-set evaluation. \lmmsevallite~ prunes unnecessary data instances to reduce the evaluation cost while maintaining the evaluation quality.

\textbf{(3)} \livebench: an evaluation benchmark that gathers the latest information from news and forum websites to evaluate models' zero-shot generalization ability on the most recent events. \livebench~ aims to provide a low-cost and generalizable way to evaluate multimodal models.

\section{\lmmseval: A Unified Multimodal Models Evaluation Suite}


 Evaluation has often taken a significant amount of time in the model development cycle. In \Cref{sec:scale_up_eval} we argue that existing evaluation pipelines in LMM contain much overhead and are not standardized. By introducing \lmmseval, we reduce this overhead and scale up the evaluation. However, as we note in \Cref{sec:trillema}, there is still a trilemma in LMM evaluation that we cannot fully resolve but only find a better trade-off.

\subsection{Scaling Evaluations with a Standardized Framework}
\label{sec:scale_up_eval}

\begin{table*}[htp]
    \centering
    \vspace{-2mm}
    \setlength{\tabcolsep}{2pt}
\renewcommand{\arraystretch}{1.1}
\resizebox{\textwidth}{!}{
\begin{tabular}{@{}l|c|cccccccc@{}}
\toprule
\textbf{Models} & \textbf{Parameters} & \textbf{AI2D} & \textbf{ChartQA} & \textbf{DocVQA} & \textbf{LLaVA$^{W}$} & \textbf{Mathvista} & \textbf{MME} & \textbf{MMMU} & \textbf{RealworldQA} \\ \midrule
LLaVA-1.5-7B & 7B & 54.8 & 18.2 & 28.1 & 59.6 & 26.7 & 1859.0 & 35.3 & 55.8 \\
LLaVA-NeXT-Vicuna-7B & 7B & 66.6 & 54.8 & 74.4 & 72.3 & 34.4 & 1841.8 & 35.1 & 57.8 \\
LLaVA-NeXT-Mistral-7B & 7B & 60.8 & 38.8 & 72.2 & 71.7 & 37.4 & 1823.4 & 33.4 & 59.3 \\
Qwen-VL-Chat & 7B & 45.9 & 60.1 & 66.3 & 21.2 & 24.6 & 1890.8 & 27.7 & 1.7 \\
InstructBLIP-Vicuna-7B & 7B & 33.8 & 12.5 & 13.9 & 55.2 & 23.4 & 1508.7 & 28.4 & 37.4 \\
LLaVA-NeXT-LLaMA3-8B & 8B & 71.6 & 69.5 & 78.2 & 80.1 & 37.5 & 1971.5 & 41.7 & 60.0 \\
Xcomposer4K-HD & 8B & 78.1 & 80.6 & 90.8 & 74.2 & 57.3 & 2189.8 & 42.6 & 62.6 \\
Idefics2-8B & 8B & 69.2 & 26.4 & 73.4 & 43.7 & 48.0 & 1792.1 & 39.7 & 25.5 \\
LLaVA-1.5-13B & 13B & 59.5 & 18.2 & 30.3 & 66.1 & 26.4 & 1818.3 & 34.8 & 54.9 \\
LLaVA-NeXT-Vicuna-13B & 13B & 70.0 & 62.2 & 77.5 & 72.3 & 35.1 & 1891.9 & 35.9 & 58.7 \\
InstructBLIP-Vicuna-13B & 13B & 36.8 & 12.7 & 13.6 & 54.4 & 25.0 & 1529.6 & 33.7 & 42.4 \\
InternVL-1.5 & 26B & 79.0 & 83.8 & 92.4 & 90.2 & 61.5 & 2183.6 & 43.1 & 65.0 \\
LLaVA-NeXT-34B & 34B & 74.9 & 68.7 & 84.0 & 88.8 & 46.0 & 2030.4 & 46.7 & 62.0 \\
LLaVA-NeXT-72B & 72B & 77.4 & 77.0 & 84.4 & 89.2 & 46.6 & 2158.9 & 46.4 & 65.4 \\
LLaVA-NeXT-110B & 110B & 80.4 & 79.7 & 85.7 & 90.4 & 49.0 & 2200.4 & 49.1 & 63.1 \\ 
LLaVA-OV-0.5B     & 0.5B & 57.1 & 61.4 & 73.7 & 74.2 & 34.8 & 1478.0 & 31.4 & 55.6 \\
LLaVA-OV-0.5B(SI) & 0.5B & 54.2 & 61.0 & 75.0 & 71.2 & 34.6 & 1489.0 & 31.2 & 53.7 \\
LLaVA-OV-7B       & 7B   & 81.4 & 80   & 90.2 & 90.7 & 63.2 & 1998.0 & 48.8 & 66.3 \\
LLaVA-OV-7B(SI)   & 7B   & 81.6 & 78.8 & 89.3 & 86.9 & 56.1 & 2109.0 & 47.3 & 65.5 \\
LLaVA-OV-72B      & 72B  & \textbf{85.6} & 83.7 & 93.1 & 93.5 & \textbf{67.5} & 2261.0 & 56.8 & 71.9 \\
LLaVA-OV-72B(SI)  & 72B  & 85.1 & \textbf{84.9} & \textbf{93.5} & \textbf{93.7} & 66.5 & \textbf{2269.0} & \textbf{57.4} & \textbf{73.8} \\
\bottomrule
\end{tabular}%
}

    \caption{An overview of selected results on \lmmseval, achieved through a standardized and transparently reproducible pipeline.}
    \label{tab:fair-comparison}
    \vspace{-10pt}
\end{table*}

\paragraph{Reducing the overhead} Existing evaluations in LMMs are often done on a model-by-model and dataset-by-dataset basis~\citep{liu2023visual, qwenvl2024}. Researchers create custom inference scripts for their models across different benchmarks. While manageable for a single model and a few benchmarks, this process becomes highly inefficient when evaluating multiple checkpoints across ten or more datasets. Users need to manually launch each individual script to preprocess the datasets, inference models, and calculate final scores based on the outputs. Boilerplates are also abundant in the code. To address this, \lmmseval~follows the framework design of \textsc{lm-eval-harness}~\citep{eval-harness} to allow for a one-command evaluation of multiple models and datasets. We preprocess and handle all the data needed during evaluation, ensuring a single data source is used across different models for a standardized evaluation. Furthermore, detailed model outputs and results will be logged for future analysis.

\paragraph{Standardized evaluation} Custom evaluation scripts also lead to another issue: the scores reported in different places are not directly comparable. For instance, \citep{li2023seedbench} extracts model answers by comparing the output probabilities among the choices. It is counted correct so long as the ground-truth answer has the lowest perplexity among the choices (PPL-based). However, \citep{liu2023improvedllava} use the generation-based evaluation. An answer is counted as correct only if the model's generation matches the option letter. To this end, we design a unified framework in \lmmseval \space covering different evaluation setups. We believe there is no best setup but one needs to fix one when comparing results across different models. For a fair comparison, we also respect the chat template of the models if they are instruction-tuned. For reproducibility and transparency, a detailed log containing the evaluation setup, model generations, and score breakdown will be automatically logged. Since we designed a unified interface, new models and datasets can also be quickly added into \lmmseval.

Equipped with these two core designs, we successfully scaled up our evaluation to over 10 models and more than 50 datasets. We present partial results in \Cref{tab:fair-comparison} and the full supported models, datasets, and scores can be found in \Cref{appendix:suite-info} and \Cref{appendix:lmms-eval-more}. We believe that large-scale evaluations are crucial. They enable a comprehensive comparison across various aspects of model performance, revealing whether a model is a versatile performer or excels only in specific tasks.  Additionally, large-scale, reproducible, and standardized evaluations are essential in ablation experiments to enhance our understanding of model architectures and training data.

\subsection{The Evaluation Trilemma}
\label{sec:trillema}
Our ultimate goal is to find a wide-coverage, low-cost, and zero-contamination way to evaluate LMMs. However, even with \lmmseval, we find it to be hard or even impossible. Specifically, once we scale the evaluation datasets to 50+, it becomes time-consuming to perform a full evaluation run on those datasets. Besides, those benchmarks are also susceptible to contamination during the training time\citep{yang2023rethinking}. As shown in Figure \ref{fig:lmms_eval_banner}, we believe there is a trilemma in model evaluation. One can not achieve the three goals simultaneously but only find a trade-off. The LMSys Chatbot Arena~\citep{chiang2024chatbot}and AI2 WildVision~\citep{yujie2024wildvisionarena} are foundational works in stressing wide coverage and anti-contamination. We present our solution to balance the other two sides of the triangle in \Cref{sec:evallite} and \Cref{sec:livebench}.

\section{\lmmsevallite: Affordable Evaluation with Broad Domain Coverage}
\label{sec:evallite}

\begin{figure}
    \centering
    \vspace{-0.2cm}
    \includegraphics[width=0.45\textwidth]{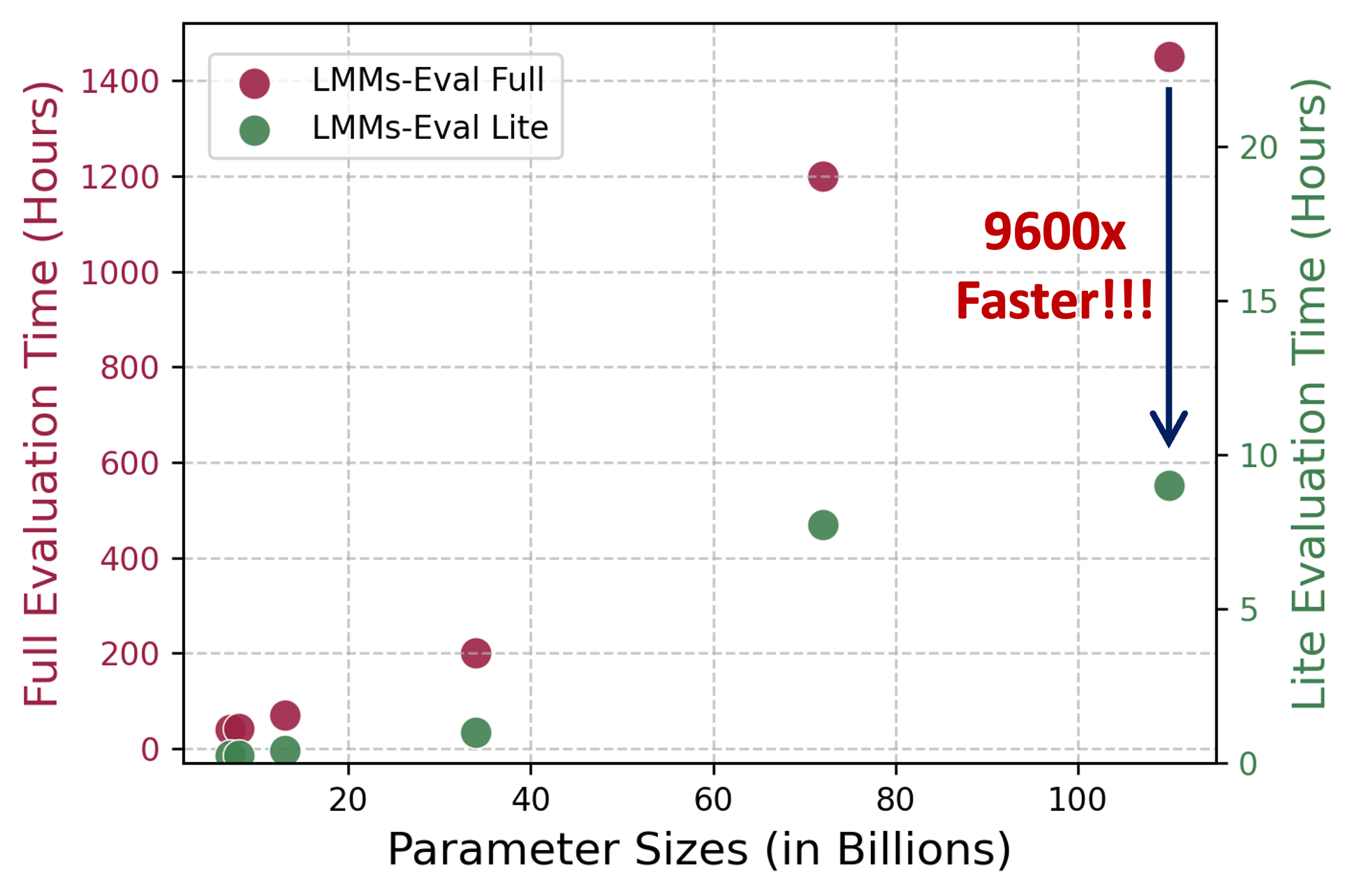}
    \caption{Evaluation cost demonstration on \texttt{Full} and \texttt{Lite} set.} 
    \vspace{-0.2cm}
    \label{fig:cost}
    \vspace{-0.2cm}
\end{figure}

 We estimate the time to evaluate various LLaVA models on all \lmmseval~datasets in Figure \ref{fig:cost}. These evaluations were conducted using 8×A100 GPUs with flash attention enabled. We replicate the model weights across GPUs and use data parallel by default. For models larger than 72B, we use pipeline parallelism~\citep{huang2019gpipe} to load a single model across different GPUs. 

We aim to construct a lite benchmark set that can provide useful and fast signals during the model development. If we can identify a subset of the benchmark where the absolute scores and relative rankings among models remain similar to the full set, we can consider it to be safe to prune the datasets. We thus present \lmmsevallite~ to complement the full datasets in \lmmseval.


%

\paragraph{Lite set selection} Let the benchmark be represented as \(D = \{ (x_{i}, y_{i}) \}_{i=1}^{n}\) and the scoring function underlying the benchmark system be denoted as \(S\). Given a model \(f\), let the response of the model to a particular question in the dataset be denoted as \(f(x_{i}) = \widehat{y}_{i}\). We aim to select a subset of the benchmark \(V \in D\) such that

\begin{align*}
    \label{eq:min_loss} &
    \min_{V:\left | V \right | \leq \left | D \right |} \left | \frac{1}{\left | D \right |} \sum_{i=1}^{\left | D \right |} S(y_{i}, \widehat{y}_{i}) -  \frac{1}{\left | V \right |} \sum_{i=1}^{\left | V \right |} S(y_{i}, \widehat{y}_{i}) \right |
\end{align*}
This objective function is equivalent to solving the $k$-Center problem \citep{sener2018active}, which seeks to identify a subset of data points that represent the full set. Thus, our problem is reformulated as finding representative points in $x_{i}$, which has been proven to be solvable as a $k$-Center problem \citep{sener2018active}. Since solving the $k$-Center problem is NP-hard \citep{cook1997combinatorial}, we use a greedy algorithm to achieve a \(2\)-OPT solution efficiently (details in ~\Cref{appendix:greedy}).

For $k$-center clustering, embeddings are extracted for each data point. While \citep{sener2018active} used CNN for image embeddings, we employed CLIP \citep{radford2021learning} for image embeddings and BGE-M3 \citep{bge-m3} for text embeddings, concatenating them to form the final embedding.

\begin{table}[h]
    \centering
    \centering
    \resizebox{0.48\textwidth}{!}{%
    \setlength{\tabcolsep}{12pt}
    \renewcommand{\arraystretch}{1.0}
    \begin{tabular}{l|llll}
    \toprule
    Dataset   & Quire & $k$-means & Lite(Ours) &  \\
    \midrule
    Flickr30k & 0.97         & 0.79    & 0.91 &  \\
    AI2D      & 0.45         & 0.87    & \textbf{0.98} &  \\
    SeedBench & 0.27         & 0.87    & \textbf{0.87} &  \\
    TextVQA   & 0.99         & 0.98    & \textbf{0.99} & \\
    \bottomrule
    \end{tabular}
    }
    \caption{Correlation results on multiple benchmarks and comparisons with $k$-means~\citep{lloyd1982least} and Quire ~\citep{quire}}
    \label{tab:correlation}
\end{table}

\begin{figure}[h]
    \centering
    \begin{subfigure}{0.45\textwidth}
        \centering
        \includegraphics[width=\textwidth]{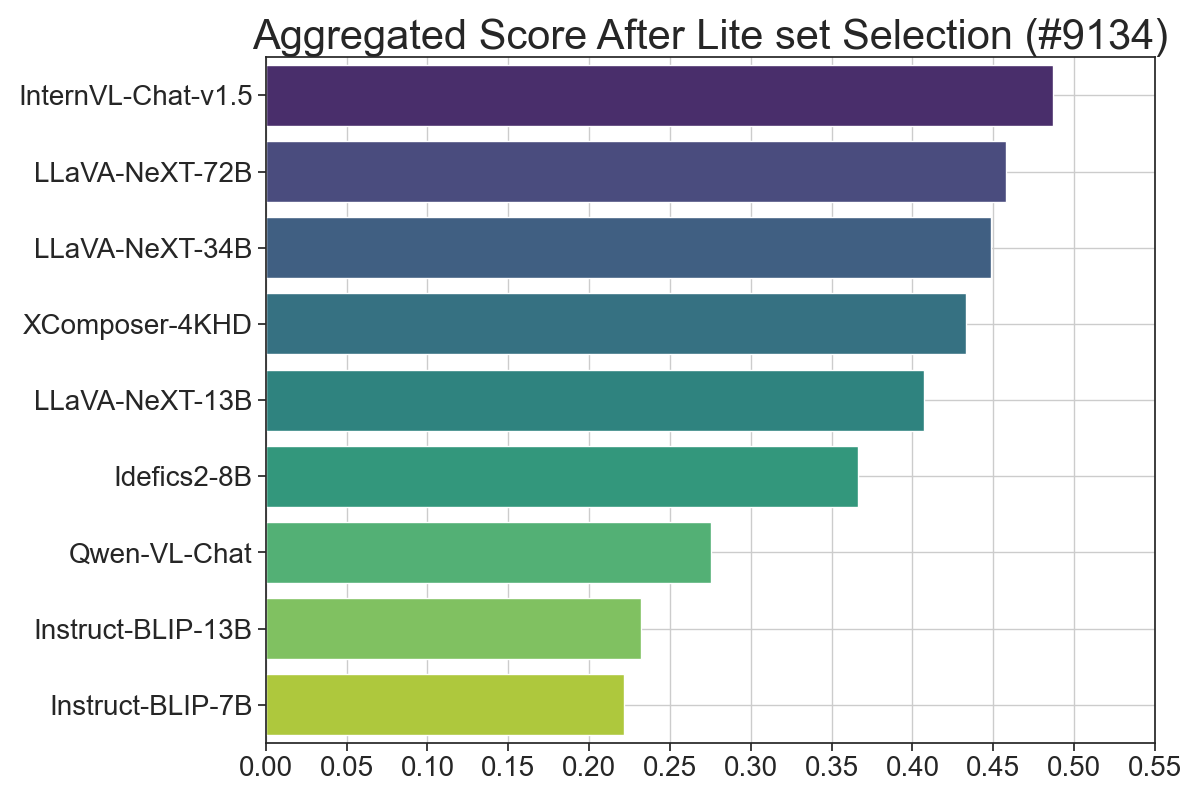}
    \end{subfigure}
    \hfill  
    \begin{subfigure}{0.45\textwidth}
        \centering
        \includegraphics[width=\textwidth]{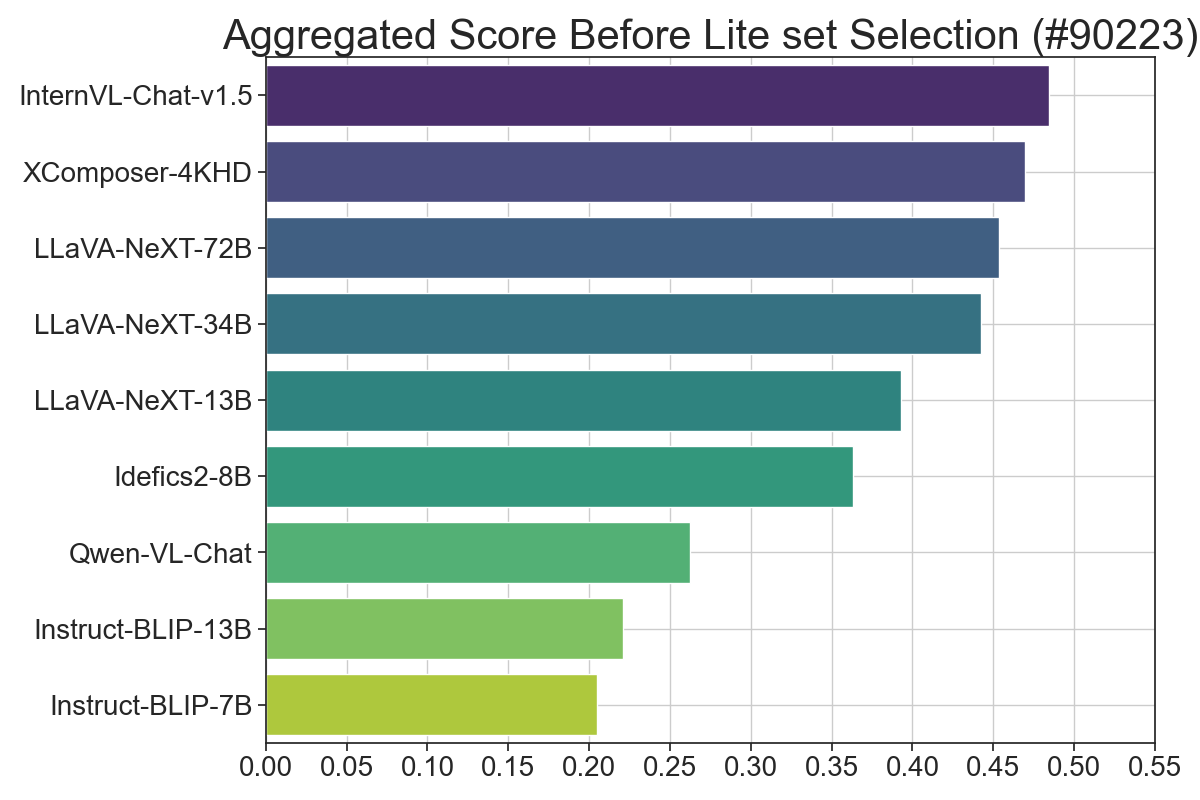}
    \end{subfigure}
    \caption{Results of \lmmsevallite \space across different models. The $x$-axis represent the weighted average percentage of scores that the model get across all the dataset.}
    \label{fig:lmms-eval-lite-results}
\end{figure}

To ensure our selected subset retains basic testing abilities compared to the original benchmarks, we assess the correlation between the original scores and the lite set scores across six versions of LLaVA \citep{liu2023improvedllava}. As shown in \Cref{tab:correlation}, our method maintain decent correlation results. Since the application of coreset selection in evaluating LMM datasets is limited and we are among the first to explore this approach to the best of our knowledge. There are only few methods for comparison. Additional results are provided in ~\Cref{appendix:lmms-lite-plus}.

\paragraph{Lite benchmark construction} We refer to datasets from works like \citep{2023GPT4VisionSC,geminiteam2024gemini,claude3family2024,liu2023improvedllava} to construct \lmmsevallite, selecting 15 datasets across different task domains for broad coverage. To keep evaluation costs low, we apply a selection method to choose representative points from datasets with over 1500 data points. For MME \citep{fu2024mme}, due to low correlation between the original and lite set scores, we retain the full version. In addition, we curate a new version of \lmmsevallite~ in \Cref{appendix:lmms-lite-plus} that contains more datasets.

%

\paragraph{Score Aggregation} To provide an overall signal to guide model development, we designed a strategy to aggregate the scores across different benchmarks in \lmmsevallite. Since different datasets and benchmarks come up with their own metrics, it is not reasonable to simply calculate the average score. Instead, we first normalize the scores from each dataset within a range of 100 and then calculate the average to be the final aggregated score. We report the aggregated score before and after the lite set pruning in \Cref{fig:lmms-eval-lite-results} to demonstrate the effectiveness of our selection method. Note that \lmmsevallite~is not designed to fully compare the performance of different model families. Instead, it served as a tool to provide useful and low-cost signals during model training and ablations.

\section{\livebench: From Static to Live Evaluation}
\label{sec:livebench}
\subsection{Probing into Multimodal Data Contamination}
\label{sec:decontainmination}

LMMs are trained on massive amounts of data.
For instance, Qwen-VL \citep{bai2023qwenvl} leverages 1.4 billion pretraining data and CogVLM \citep{wang2024cogvlm} uses 1.5 billion. 
However, research in both LLMs \citep{zhang2024careful,wei2023skywork} and LMMs \citep{chen2024mmstar} has indicated that data contamination can significantly skew benchmark scores. This highlights the need for careful data management and validation to ensure accurate and fair evaluations.

We explore multimodal training within the LLaVA frameworks, utilizing two primary data types: (1) pretraining data to align visual and textual embeddings and train the vision encoder, and (2) high-quality, supervised finetuning data to improve diverse instruction-following capabilities. 
The re-annotation and conversion of large web and academic datasets into training materials frequently lead to issues of overlap and contamination. To address this, we developed an analytical tool to assess the overlap between training and benchmark data, showcasing our findings with data from \citep{liu2023improvedllava} with user data removed in it.



\begin{figure}[ht]
    \centering
    \begin{subfigure}{0.45\textwidth}
        \centering
        \includegraphics[width=\textwidth]{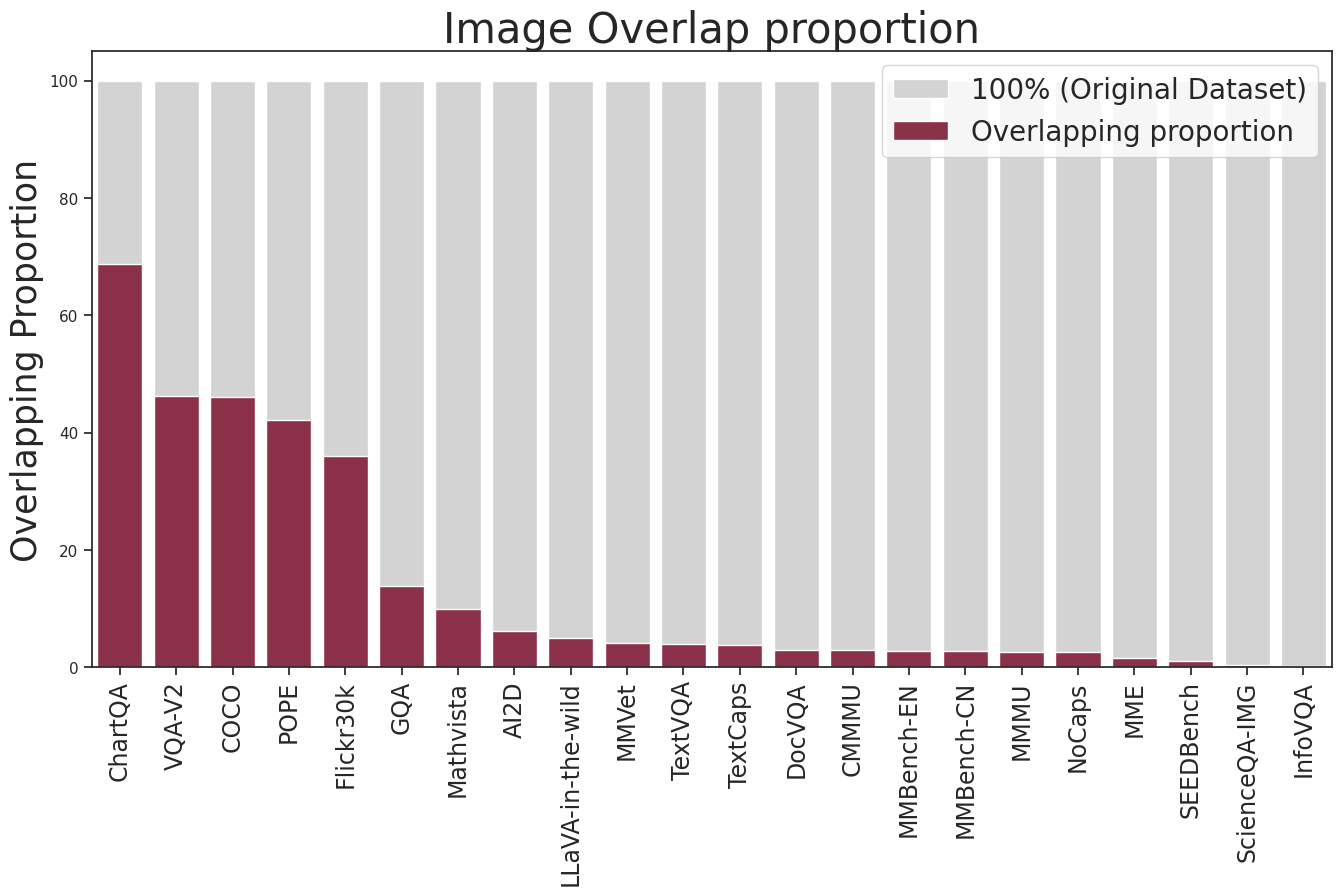}
    \end{subfigure}
    \hfill  
    \begin{subfigure}{0.45\textwidth}
        \centering
        \includegraphics[width=\textwidth]{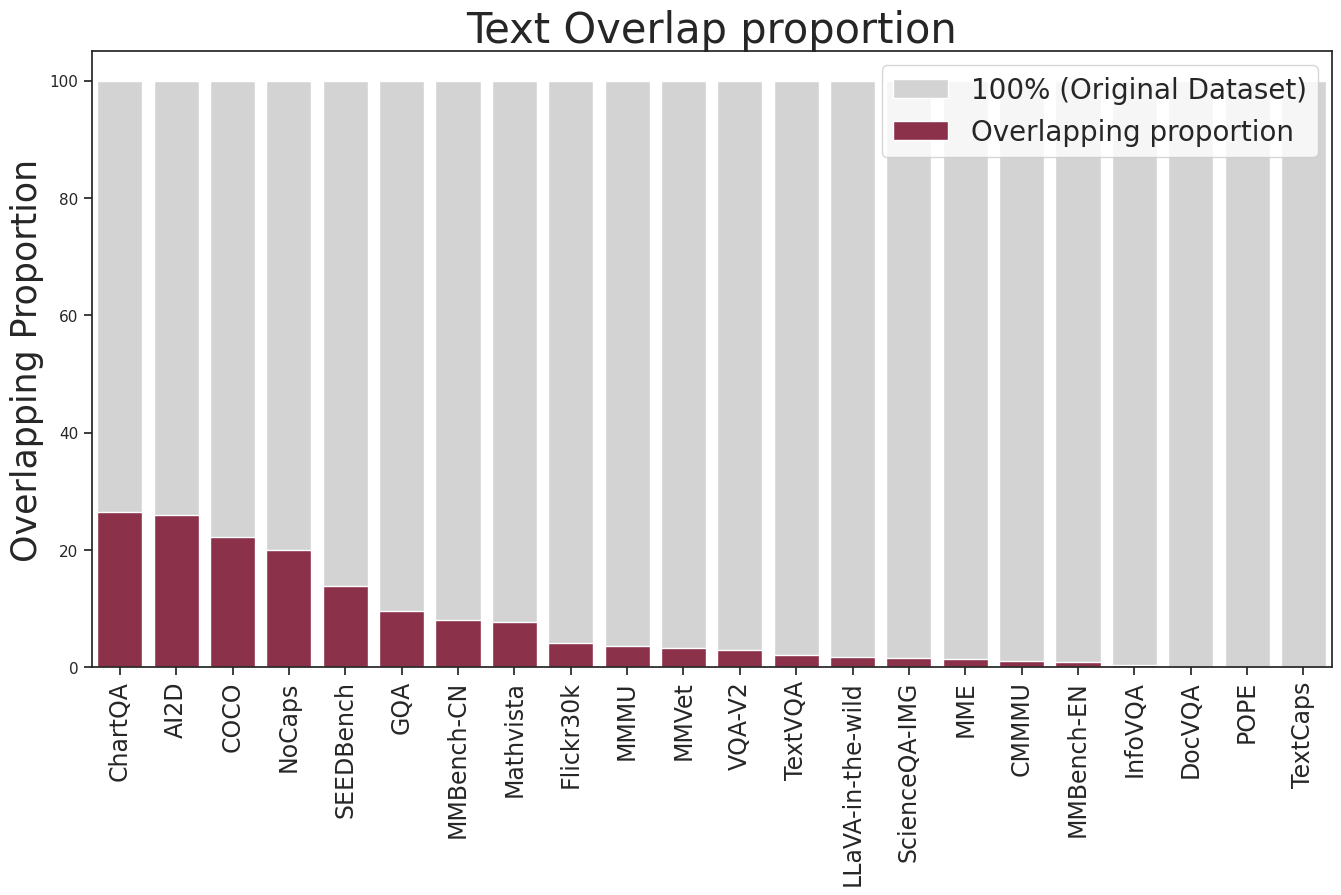}
    \end{subfigure}
    \caption{Contamination analysis in current evaluation benchmarks and LLaVA's training data. Among the datasets with an overlap proportion exceeding 20\%, including ChartQA, VQAv2, COCO2014, and GQA, it has been confirmed that their training sets are included in LLaVA's training data.}
    \label{fig:Contamination proportion}
    \vspace{-2mm}
\end{figure}

\begin{figure*}[t]
    \centering
    \includegraphics[width=1.0\textwidth]{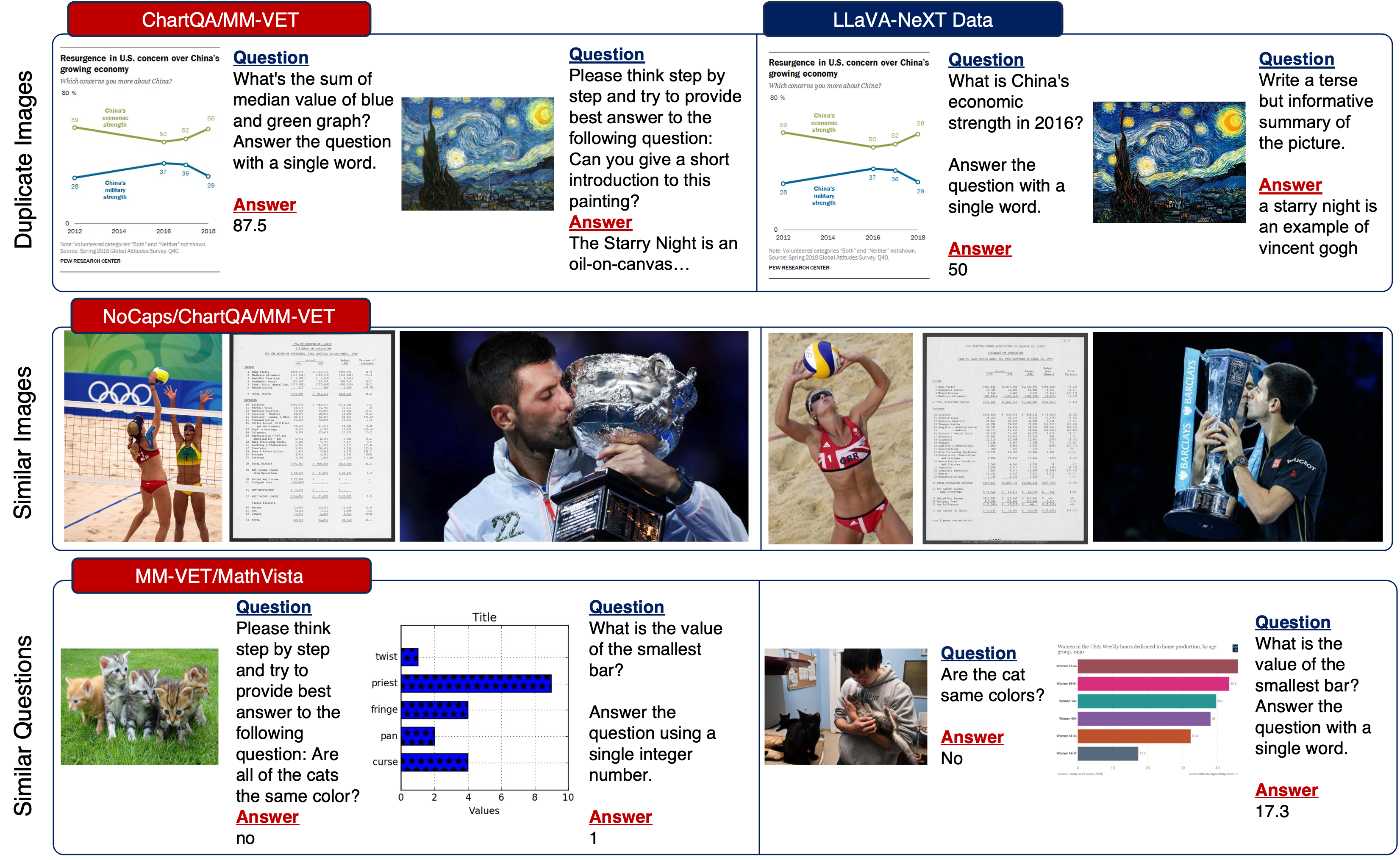}
    \caption{We present several cases of possible data overlapping in LLaVA-NeXT pretraining and supervised-finetuning data. We observed three types of data contamination (1) duplicate images (2) similar images (3) similar questions.}
    \label{fig:qualitative}
    \vspace{-0.5cm}
\end{figure*}


\paragraph{Text Overlap} To measure text overlap, we use a string matching technique similar to those by GPT-4 \citep{openai2024gpt4}, PaLM \citep{anil2023palm}, and LLaMA \citep{touvron2023llama}. Typically, an $8\sim13$ n-grams range is used \citep{brown2020language}, but we consistently use $8$ n-grams for simiplicity. We exclude any n-gram appearing more than $10$ times in the training data, labeling these as \textit{meaningless n-grams}. We also calculate an overlap ratio for each new n-gram candidate against our set of meaningless n-grams, excluding those exceeding a predefined threshold.

\paragraph{Image Overlap} Contrary to text overlap, determining image overlap is a more challenging task. While it is common practice to compute image embeddings and then calculate their cosine similarity, selecting an appropriate threshold applicable to all datasets is difficult. Instead of computing similarity in the embedding space, we empirically find that using the pretrained SEED-tokenizer~\citep{ge2023making} leads to meaningful separation in detecting the overlap. We first tokenize each image into a 1-D sequence of 32 tokens. Similar to text, an 8-gram lookup table was constructed from those image tokens to detect image contamination. The occurrence of 8-gram overlap can be interpreted as approximately $1/4$ of the image overlapping.

\subsubsection{Results \& Analysis on Decontamination}

To evaluate the potential contamination of current benchmarks, we selected over 20 benchmarks, including AI2D \citep{kembhavi2016diagram}, ChartQA \citep{masry2022chartqa}, NoCaps \citep{agrawal2019nocaps}, VQA v2 \citep{balanced_vqa_v2}, and LLaVA-in-the-wild \citep{liu2023visual}. We report the percentages of image and text overlap in ~\Cref{fig:Contamination proportion} for our selected datasets and more qualitative results qualitative results in Figure \ref{fig:qualitative}. Our examination of both image and text overlaps has revealed three primary types of data contamination across various benchmarks.

\paragraph{Duplicate Images} Instances of completely identical images between the training set and benchmark datasets were observed. This issue is exemplified by two identical images in ChartQA \citep{masry2022chartqa} and MM-Vet \citep{yu2023mmvet}. 

\paragraph{Similar Images} Our image n-gram analysis has succesfully identified the occurrence of visually similar images in both the training and benchmark datasets. Such similarities could lead to semantically similar questions, as demonstrated in examples from NoCaps~\citep{agrawal2019nocaps}, ChartQA~\citep{masry2022chartqa} and MM-Vet~\citep{yu2023mmvet}.

\paragraph{Similar Questions} We also observe recurring question structures in the training data that mirror those in the benchmark dataset. Although the corresponding images may differ, the similarity in question structure could advantage the model in responding to benchmark queries.

\begin{figure*}[htp]
    \centering
    \includegraphics[width=1\textwidth]{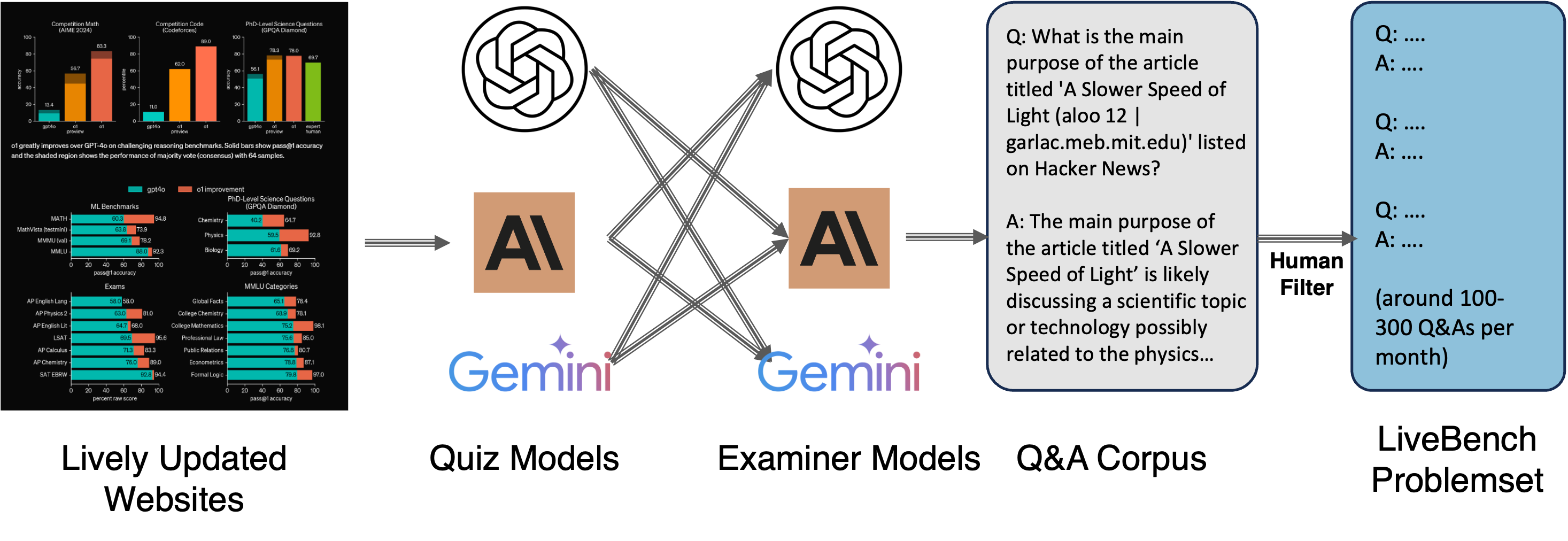}
    \caption{Overview pipeline for \livebench~. We collect the latest information from the lively updated websites, organize the Q\&A based on the information with the assistance of multimodal models, verify the Q\&A with human annotators, evaluate the models with the Q\&A corpus using different judge models, including human judges, and finally report the problemset.}
    \label{fig:livebench}
\end{figure*}

\subsection{Multimodal LiveBench}

Traditional benchmarks rely on static evaluations with fixed questions and answers. While open-source models often outperform commercial ones like GPT-4V in benchmarks, they fall short in real user experience. Dynamic, user-oriented arenas like LMSys and WildVision are gaining popularity but face issues with prompt quality, difficulty, and noisy traffic, making consistent comparisons tough and costly. New benchmarks like Vibe-Eval~\citep{padlewski2024vibe} and LLaVA-Wilder~\citep{li2024llavanext-strong} use real-world data for more authentic testing, but as models continuously update from web data, there’s a risk of contamination in evaluation benchmarks.

We propose \livebench, a new evaluation framework that uses a dynamically updated dataset to prevent contamination and reduce costs. The evaluation data is collected from webpages, with an automated pipeline that gathers the latest global information from sources like news sites and community forums.

\subsubsection{Dataset Curation Process}
\paragraph{Data Collection From the Web} To ensure the timeliness and authenticity of our information, we select sources from over 60 news outlets, including CNN, BBC, Japan's Asahi Shimbun, and China's Xinhua News Agency, as well as insights from forums like Reddit. A detailed list of these sources is provided in~\cref{appendix:website_list}. 

\paragraph{Information Extraction} The data collection pipeline is illustrated in~\cref{fig:livebench}, where the process begins by capturing screenshots of news website homepages. The information extraction consists of three main steps. \textbf{1)} First, the model performs OCR to extract all text from the website. \textbf{2)} The model is then instructed to identify significant images within the screenshot and extract relevant details about these images, such as the environment depicted, the actions and expressions of individuals, and the relationship between the images and the corresponding text. \textbf{3)} Finally, the model is asked to specify what makes the information "newsworthy." For example, if the news is about the U.S. election, the model identifies what occurred in September 2024 that differentiates this news. Throughout the extraction process, we use Claude-3.5-Sonnet. All the prompts in this process can be found in ~\cref{tab:info-extractor}

\paragraph{QA Generation} The extracted information is then sent to the quiz model to generate questions and answers (QA). The model is prompted to create questions for four categories: \textbf{(1)} Concrete Recognition, \textbf{(2)} Real-world Application, \textbf{(3)} Analytical Understanding, and \textbf{(4)} Divergent Thinking \& Creation. These categories are based on Bloom's Taxonomy~\citep{bloom1956taxonomy}. We prompt the model to produce challenging and innovative questions, along with criteria for scoring them. Detailed explanations of these categories and the prompts used to generate QA are provided in ~\cref{tab:qa-generator}. An example QA with criteria can be found in ~\cref{tab:question_criteria}

\paragraph{QA Checker \& Finalizer} To further curate high-quality QA pairs, we introduce the Checker and Finalizer models to refine the details of the QA pairs and validate the answers. The Checker model is mainly responsible for refining the questions and answers, restructuring them to ensure the questions are more answerable, verifiable, and challenging. It also ensures that the QA falls into the correct category. If the QA does not meet the requirements, the Checker model modifies the question and forwards it to the Finalizer. The Finalizer is mainly responsible for reformatting the question to enhance readability for human users. The prompt we use is included in ~\cref{tab:finalizer,tab:checker}

\paragraph{QA Scorer} The final part of our pipeline involves a scorer, which evaluates the QA pairs based on three criteria: Authenticity, Logical Coherence, and Clarity and Precision, assigning a score from 1 to 10. To balance data collection costs with evaluation efficiency, we collect approximately 500 questions each month and select 100 to 300 for the final \livebench~problem set, based on those that exceed a certain score threshold. We also manually review the questions to remove any that are inappropriate. You can find the prompt in ~\cref{tab:scorer}

We provide 4 examples for each category in ~\cref{tab:question_understanding,tab:question_creation,tab:question_analytical,tab:question_application}. It is important to note that the quality of our QA may still fall below that of human-curated answers, as we are aiming to build a dynamic evaluation pipeline that strikes a balance between cost and broad coverage.

%
%

\subsubsection{Evaluation Metrics \& Results on \livebench}

We adopt the scoring criteria from LLaVA-Wilder~\citep{li2024llavanext-strong} and Vibe-Eval~\citep{padlewski2024vibe}, using GPT-4o as the primary judge model. The judge assigns scores from $1$ to $10$ based on ground-truth answers and the scoring criteria. By leveraging established criteria, our evaluations are comprehensive and aligned with current standards. Detailed criteria and evaluation prompts are provided in~\cref{appendix:evalution_prompts}.

\begin{table}[htp]
    \centering
    \setlength{\tabcolsep}{4pt}
    \resizebox{0.5\textwidth}{!}{
        \begin{tabular}{c|c|cccc}
        \toprule
        \textbf{Model} & \textbf{Overall} & \textbf{Recognition} & \textbf{Analysis} & \textbf{Thinking} & \textbf{Realworld} \\ \midrule
        LLaVA-1.5-7B             & 30.2 & 9.4   & 36.4 & 45.4 & 29.4  \\
        LLaVA-OV-0.5B            & 32.4 & 25.1  & 33.6 & 40.2 & 30.6  \\
        LLaVA-OV-7B              & 64.9 & 57.2  & 67.0 & 76.2 & 59.0  \\
        LLaVA-OV-7B-Chat         & 65.6 & 48.8  & 75.8 & 84.0 & 53.6  \\
        LLaMA-3.2-V-11B-Instruct & 65.8 & 51.9  & 65.2 & 71.4 & 74.7  \\
        InternVL2-8B             & 69.6 & 65.6  & 74.8 & 77.5 & 60.4  \\
        LLaVA-OV-72B-Chat        & 75.0 & 62.0  & 87.8 & 83.8 & 66.6  \\
        Qwen2-VL-7B              & 79.2 & 74.2  & 82.8 & 87.4 & 75.2  \\
        Gemini-1.5-Flash         & 81.6 & 77.1  & 82.4 & 89.0 & 77.9  \\
        Gemini-1.5-Pro           & 84.5 & 85.4  & 83.8 & 88.6 & 80.1  \\
        Qwen2-VL-72B             & 85.9 & 86.7  & 88.8 & 89.0 & 79.2  \\
        Claude-3.5-sonnet        & 90.3 & 94.6  & 93.4 & 95.3 & 85.8  \\
        GPT4o-mini              & 91.9 & \textbf{94.6}  & 93.4 & \textbf{95.3} & 84.3  \\ 
        GPT4o                   & \textbf{92.0} & 91.7  & \textbf{93.8} & 94.8 & \textbf{87.6}  \\ \bottomrule
        \end{tabular}%
        }
    \caption{\texttt{LiveBench-2024-09} Results.}
    \label{tab:livebench_results}    
\end{table}

The results in~\Cref{tab:livebench_results} indicate that the GPT-4 series models, including GPT-4o-mini and GPT-4o, are among the top performers, whereas the Gemini and Claude series models still outperform open-source models. GPT-4o has a large lead on recognition ability along with some small lead in other abilities. We provide a detailed case analysis in ~\cref{subsec:livebench-case-analysis} with many case studies to demonstrate how GPT-4o outperforms other models in many cases.

\paragraph{Open-sourced models are still far from achieving the level of GPT-4V.} The \textit{current superiority in benchmarks} can be attributed to the simplicity, fixed nature, or potential contamination of the evaluated scenarios (e.g., MME~\citep{fu2024mme} and MMBench~\citep{liu2024mmbench}). These observations align with our hypothesis regarding the strengths and limitations of commercial multimodal models like GPT-4V, which exhibit robust capabilities that existing benchmarks do not fully assess. 

Specifically, our \livebench~ requires models to demonstrate strong zero-shot generalization abilities, as they must interpret continuously updated content from news and forum websites, highlighting the unique advantages of these commercial models.

While these findings may appear disadvantageous for competitors, they reveal the shortcomings of traditional benchmarks and emphasize the necessity for more comprehensive evaluations to accurately assess model performance. Benchmarking remains a crucial tool for driving progress in AI, and these results provide valuable insights for future contenders aiming to enhance their models.

\section{Conclusions}
\vspace{-1mm}
In this work, we conducted a thorough reality check on the current evaluation pipeline and benchmarks for LMMs. We recognize the difficulties in the evaluation due to the \textit{evaluation trilemma.} Although we cannot break this trilemma, we present three key contributions to find a better trade-off: \textbf{1)} \lmmseval, a unified evaluation suite for a standardized and large-scale LMM evaluation, \textbf{2)} \lmmsevallite~to balance low-cost evaluation with wide coverage, and \textbf{3)} \livebench, a benchmark that transforms traditional static evaluation into a dynamic format to address potential data contamination in LMMs evaluation. We hope our \lmmseval~family makes a valuable contribution to the community towards the holistic evaluation of LMMs.

\section{Limitations}
Through reality check, we explore the field of evaluation in LMMs and re-examine the evaluation process. Throughout our papers, we assume that the evaluation trilemma cannot be resolved. This suggests future work that goes deeper into finding a better trade-off among the sides of the trilemma or potentially overcoming it. Additionally, we address the issue of data contamination using a relatively simple method that requires access to the training data, while most research does not open-source their data. Future work may focus on methods that rely solely on the model and develop more efficient approaches.

\section*{Acknowledgments}

This study is supported by the Ministry of Education, Singapore, under its MOE AcRF Tier 2 (MOE-T2EP20221-0012, MOE-T2EP20223-0002), and under the RIE2020 Industry Alignment Fund – Industry Collaboration Projects (IAF-ICP) Funding Initiative, as well as cash and in-kind contribution from the industry partner(s).

\bibliography{custom}

\appendix

\clearpage
\appendix

\section{Related Work}

\paragraph{Vision language benchmark} Historically, benchmarks such as AI2D~\citep{kembhavi2016diagram}, TextVQA~\citep{singh2019vqa}, TextCaps~\citep{sidorov2020textcaps}, Flickr30k~\citep{young-etal-2014-image}, and OK-VQA~\citep{okvqa} were used to assess computer vision model's individual performance in captioning, optical character recognition, and visual question answering. With the emergence of Large Language Models (LLMs), Large Multimodal Models (LMMs) have been developed~\citep{zhang2024mmllmsrecentadvancesmultimodal} to emphasize more comprehensive capabilities across vision and language. Subsequently, new benchmarks featuring increasingly challenging tasks and more holistic evaluation were proposed.
For instance, benchmarks like ScienceQA~\citep{lu2022learn} and MathVista~\citep{lu2024mathvista} evaluate math and science abilities. SEED-Bench~\citep{li2023seedbench}, CMMMU~\citep{zhang2024cmmmu}, MMMU~\citep{yue2023mmmu}, and MM-Bench~\citep{liu2024mmbench}, assess multiple heterogeneous dimensions of multimodal models/ In this paper, we aim to provide a comprehensive review of benchmarks from various fields.

\paragraph{Data contamination} The issue of data contamination has emerged as a significant concern in the evaluation of Large Language Models (LLMs). Studies by \citep{yang2023rethinking}, \citep{wei2023skywork}, and \citep{zhang2024careful} highlighted that data contamination poses a serious challenge for current LLMs and may lead to inaccuracies in accessing models' real capabilities. Methods for data decontamination include assessing n-gram overlap \citep{brown2020language}, removing similar embedding points from datasets \citep{shi2024detecting}, or leveraging influential functions \citep{koh2020understanding}. However, the issue of data contamination in benchmarks for LMMs remains relatively unexplored.

\paragraph{Coreset benchmark} With the development of numerous benchmarks, the demand for coreset versions across different benchmarks has become increasingly urgent. In LLM benchmarks, \citep{perlitz2024efficient} employ stratified random sampling to select questions, while \citep{vivek2024anchor} utilize the anchor points method for data point clustering. Other approaches, such as \citep{polo2024tinybenchmarks}, utilize Item Response Theory (IRT) \citep{lord1968statistical} to create embeddings for data points in benchmarks. In addition to these works, we have also investigated various active learning methods for efficiently and accurately constructing coresets. Quire \citep{quire} aims to select the most informative and representative points in the dataset, while \citep{mirzasoleiman2020coresets}, \citep{yang2023sustainable}, and \citep{sener2018active} focus on identifying coresets within the dataset.

\begin{table*}[h]
    \centering
    \caption{Detailed image overlap and text overlap statistics accross different dataset}
    \resizebox{0.92\textwidth}{!}{
\setlength{\tabcolsep}{8pt}
\renewcommand{\arraystretch}{1.0}
\begin{tabular}{lc|c|c}
    \toprule
    & & \multicolumn{1}{c|}{Image overlap (\%)} & \multicolumn{1}{c}{Text overlap (\%)} \\
    \cmidrule{3-4}
    Dataset & Split & LLaVA-NeXT Data & LLaVA-NeXT Data \\
    \midrule
    \multicolumn{4}{c}{\textit{Math \& Science}} \\
    \midrule
    AI2D~\citep{kembhavi2016diagram} & test & 6.09 & 25.97 \\
    MathVista~\citep{lu2024mathvista} & testmini & 9.90 & 7.70 \\
    ScienceQA~\citep{lu2022learn} & img & 0.35 & 1.54 \\
    \midrule
    \multicolumn{4}{c}{\textit{Doc \& Inforgraphic}} \\
    \midrule
    ChartQA~\citep{masry2022chartqa} & test & 68.64 & 26.52 \\
    DocVQA~\citep{mathew2020docvqa} & val & 36.08 & 4.06 \\
    InfoVQA~\citep{mathew2020docvqa} & test & 0.14 & 0.39 \\
    \midrule
    \multicolumn{4}{c}{\textit{Caption}} \\
    \midrule
    COCO2014~\citep{lin2014microsoft} & val & 46.05 & 22.19 \\
    Flickr30k~\citep{young-etal-2014-image} & test & 2.97 & 0.00 \\
    NoCaps~\citep{agrawal2019nocaps} & val & 2.53 & 19.98 \\
    TextCaps~\citep{sidorov2020textcaps} & val & 3.79 & 0.00 \\
    \midrule
    \multicolumn{4}{c}{\textit{VQA}} \\
    \midrule
    GQA~\citep{hudson2019gqa} & testdev-balanced & 13.91 & 9.50 \\
    TextVQA~\citep{singh2019vqa} & val & 3.90 & 2.00 \\
    VQAv2~\citep{balanced_vqa_v2} & val & 46.21 & 2.90 \\
    \midrule
    \multicolumn{4}{c}{\textit{Multi-task benchmark}} \\
    \midrule
    CMMMU \citep{zhang2024cmmmu} & val & 2.89 & 1.11 \\
    MMBench \citep{liu2024mmbench} & cn-dev & 2.77 & 0.81 \\
    MMBench \citep{liu2024mmbench} & en-dev & 2.77 & 7.97 \\
    MME \citep{fu2024mme} & test & 1.60 & 1.39 \\
    MMMU \citep{yue2023mmmu} & val & 2.67 & 3.56 \\
    MMVet \citep{yu2023mmvet} & val & 4.13 & 3.21 \\
    SEED-Bench \citep{li2023seedbench} & all & 1.11 & 13.84 \\
    \midrule
    \multicolumn{4}{c}{\textit{Others}} \\
    \midrule
    LLaVA-W \citep{liu2023visual} & test & 5.00 & 1.67 \\
    POPE \citep{li2023evaluating} & val & 42.20 & 0.00  \\
\bottomrule
\end{tabular}
}
    \label{tab:contamination-result}
\end{table*}

\section{Broader Impacts}
A comprehensive evaluation framework can help identify the limitations of existing multimodal models, preventing potential AI misuse. On the other hand, benchmarks can also introduce biases that may not reflect real-world scenarios. If the benchmarks are not representative of diverse applications and contexts, there is a risk that models optimized for these benchmarks may perform poorly in practical settings. Besides, automatic evaluations cannot replace expert human assessment in specialized fields such as medical imaging. The construction of \livebench~uses real-world data crawled from the web. It could potentially lead to concerns regarding data privacy. The benchmarks we provide are meant for research purposes only and should be used with caution.

\section{Data Contamination}
\label{appendix:overlapping-stats}

We present the details of the image overlapping in \Cref{tab:contamination-result}. Datasets such as ChartQA \citep{masry2022chartqa}, DocVQA \citep{mathew2020docvqa}, COCO \citep{lin2014microsoft}, and VQAv2 \citep{balanced_vqa_v2} were included in the LLaVA-NeXT \citep{liu2023improvedllava} training data and thus suffered the most from data contamination. Most of the benchmarks maintain a relatively low contamination proportion, with image and text overlap below 10\%. POPE \citep{li2023evaluating} was detected to have a high image overlapping ratio because it uses image sources from COCO \citep{lin2014microsoft}.

\subsection{More Qualitative Examples}
\label{appendix:qualitative}

\begin{figure*}[h]
    \centering
    \includegraphics[width=1.0\textwidth]{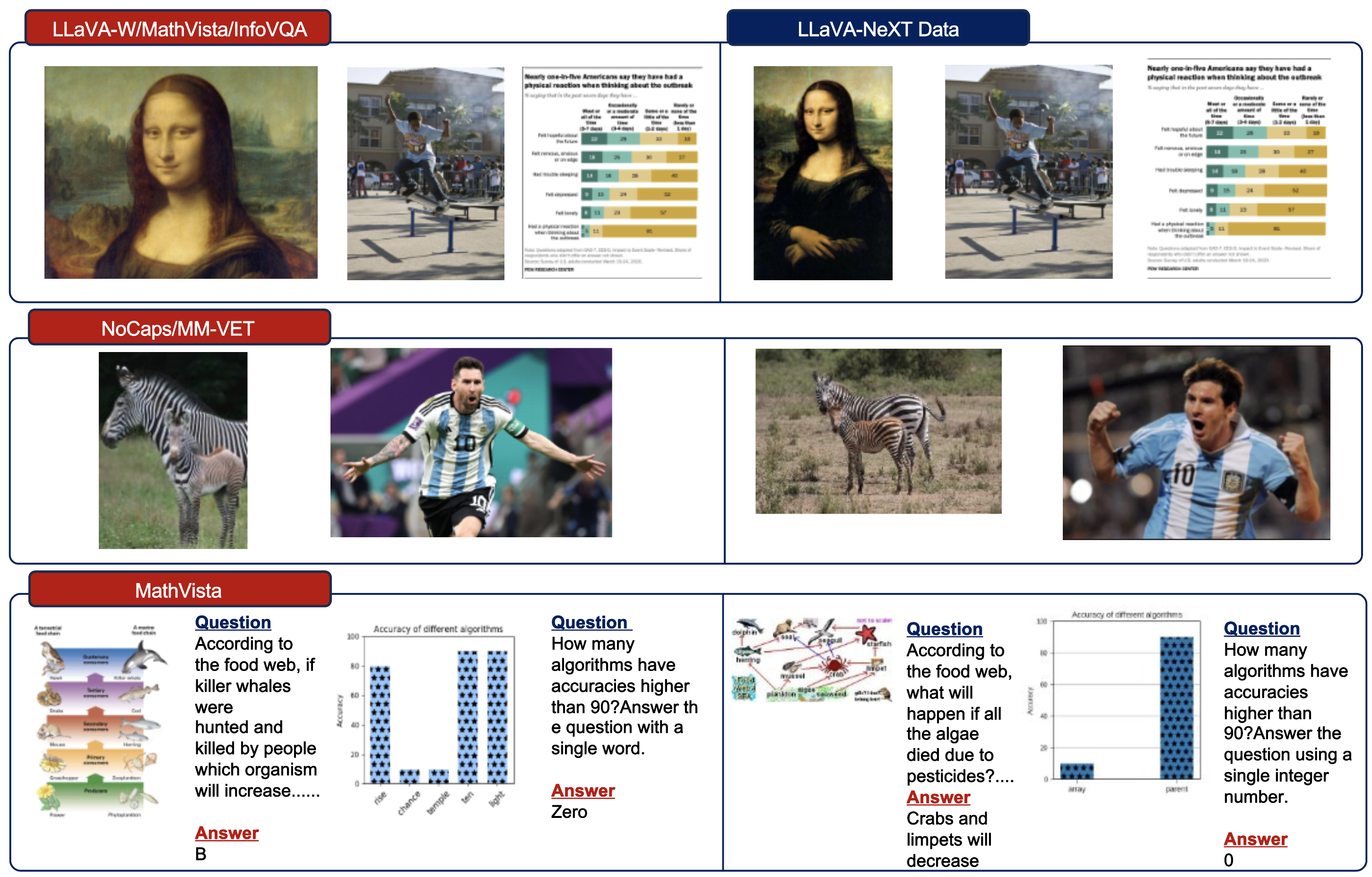}
    \caption{More qualitaive results we found using our decontamination tools}
    \label{fig:more-qualitative}
\end{figure*}

We present more qualitative results here to demonstrate the data contamination problem in the dataset. We observe more identical images in benchmarks such as LLaVA$^W$ \citep{liu2023visual}, MathVista \citep{lu2024mathvista}, and InfoVQA \citep{mathew2020docvqa}. Similar images have also been another issue in different datasets; we present two more examples in NoCaps \citep{agrawal2019nocaps} and MM-Vet \citep{yu2023mmvet}. Text overlapping can help us detect questions with similar sentence structure. Though the images might not be similar enough, these similar questions might also be marked as in-domain questions. For example, we present two cases in MathVista \citep{lu2024mathvista}. Though not necessarily contamination or overlapping cases, the two images are both testing similar domain knowledge and may help the model to answer questions in the benchmarks.

\section{~\lmmsevallite}
\subsection{Coreset Selection correlation}
\label{appendix:coreset}

\begin{table*}[h]
    \centering
    \caption{The full correlation results we achieve using our selection methods}
    \label{tab:corelation results}

\resizebox{0.85\textwidth}{!}{
    \setlength{\tabcolsep}{6pt}
    \renewcommand{\arraystretch}{1.2}
    \begin{tabular}{lc|cc|cc}
        \toprule
        & & & & \multicolumn{2}{c}{\textbf{Correlation}}\\
        \textbf{Dataset} & \textbf{Split} & \textbf{Lite Size} & \textbf{Original Size} & \multicolumn{1}{c}{LLaVA Embedding} & \multicolumn{1}{c}{CLIP+BGE Embedding} \\
        \midrule
        \multicolumn{6}{c}{\textit{Math \& Science}} \\
        \midrule
        AI2D ~\citep{kembhavi2016diagram} & test & 300 & 3088 & 0.94 & 0.98 \\
        \midrule
        \multicolumn{6}{c}{\textit{Doc \& Inforgraphic}} \\
        \midrule
        ChartQA ~\citep{masry2022chartqa} & test & 400 & 2500 & 0.96 & 0.97 \\
        DocVQA ~\citep{mathew2020docvqa} & val & 400 & 5349 & 0.99 & 0.99 \\
        InfoVQA ~\citep{mathew2020docvqa} & val & 200 & 2801 & 0.94 & 0.94 \\
        \midrule
        \multicolumn{6}{c}{\textit{Caption}} \\
        \midrule
        Flickr30k ~\citep{young-etal-2014-image}& test & 400 & 31784 & 0.99 & 0.91\\
        NoCaps ~\citep{agrawal2019nocaps}& val & 400 & 4500 & 0.99 & 0.98\\
        TextCaps ~\citep{sidorov2020textcaps} & val & 300 & 3166 & 0.98 & 0.96 \\
        RefCOCO ~\citep{kazemzadeh-etal-2014-referitgame} & val & 500 & 8811 & 0.99 & 0.99 \\
        \midrule
        \multicolumn{6}{c}{\textit{VQA}} \\
        \midrule
        TextVQA ~\citep{singh2019vqa} & val & 300 & 5000 & 0.99 & 0.99 \\
        \midrule
        \multicolumn{6}{c}{\textit{Multi-task benchmark}} \\
        \midrule
        SeedBench ~\citep{li2023seedbench} & test & 700 & 17990 & 0.77 & 0.87 \\
        \bottomrule
    \end{tabular}
}
    
\end{table*}

We compare the original scores and the selected dataset scores between the Lite version and the original datasets, calculating the correlation scores between them. We tried two different embeddings to perform $k$-center clustering. In addition to using CLIP \citep{radford2021learning} and BGE \citep{bge-m3} embeddings, we also trained a LLaVA-Qwen 1.8B model following the training recipe of \citep{liu2023improvedllava} to embed image and text pairs simultaneously. For LLaVA embeddings, the last hidden states for all tokens were averaged into a single vector to serve as the feature vector for each data point. We report the correlation results for both embeddings in \Cref{tab:corelation results}.

\subsection{Dataset statistics in \lmmsevallite~}
\label{appendix:lmms-lite}

\begin{table*}[h]
    \centering
    \caption{Overview of \lmmsevallite.}
    \label{tab:lite}

\centering
\resizebox{0.85\textwidth}{!}{%
\setlength{\tabcolsep}{12pt}
\renewcommand{\arraystretch}{1.0}
\begin{tabular}{l|l|l|l|l}
\toprule
\textbf{Task Domain} & \textbf{Dataset}    & \textbf{Split} & \textbf{Full} \textbf{Size} & \textbf{Lite} \textbf{Size} \\ \midrule
\multirow{3}{*}{Doc \& Infographic Understanding}                          & ChartQA   & test     & 2500  & 400  \\
            & DocVQA     & val   & 5349  & 400  \\
            & InfoVQA    & val   & 2801  & 200  \\ \midrule
\multirow{4}{*}{Image Understanding \& Captioning}                   & Flickr30k & val      & 31784 & 400  \\
            & NoCaps     & val   & 4500  & 400  \\
            & TextCaps   & val   & 3166  & 300  \\
            & RefCOCO    & val   & 8811  & 500  \\ \midrule
Visual Question Answering                                                  & TextVQA   & val      & 5000  & 300  \\ \midrule
\multirow{2}{*}{Math \& Science} & MathVista & testmini & 1000  & 1000 \\
            & AI2D       & test  & 3088  & 300  \\ \midrule
Visual Dialogue & LLaVA-W    & test  & 60    & 60   \\ \midrule
\multirow{4}{*}{Multi-discipline}                                          & MME       & cog. \& percep.   & 2374  & 2374 \\
            & MMMU       & val   & 900   & 900  \\
            & CMMMU      & val   & 900   & 900  \\
            & Seed-Bench & test  & 17990 & 700  \\
\midrule
\cellcolor{gray!25} - & \textbf{Total} & \cellcolor{gray!25} - & \textbf{90223} & \textbf{9134} \\ \bottomrule
\end{tabular}%
}

    \vspace{-2mm} 
\end{table*}

\begin{table*}[h]
    \centering
    \caption{\lmmsevallite~ with more datasets}
    \label{tab:lite-more}
    \centering
\resizebox{0.85\textwidth}{!}{%
\setlength{\tabcolsep}{12pt}
\renewcommand{\arraystretch}{1.0}
\begin{tabular}{l|l|l|l|l}
\toprule
\textbf{Task Domain} & \textbf{Dataset}    & \textbf{Split} & \textbf{Full} \textbf{Size} & \textbf{Lite} \textbf{Size} \\ \midrule

\multirow{3}{*}{Doc \& Infographic Understanding}                          & ChartQA   & test     & 2500  & 500  \\
            & DocVQA     & val   & 5349  & 500  \\
            & InfoVQA    & val   & 2801  & 500  \\ \midrule

\multirow{5}{*}{Image Understanding \& Captioning}                   & Flickr30k & val      & 31784 & 500  \\
            & NoCaps     & val   & 4500  & 500  \\
            & TextCaps   & val   & 3166  & 500  \\
            & RefCOCO    & val   & 8811  & 500  \\ 
            & COCO     & val   & 5000  & 500  \\
 \midrule
\multirow{5}{*}{Visual Question Answering}  & GQA   & test      & 12578  & 500  \\
    & OKVQA & val & 5046 & 500 \\
    & VizWiz-VQA & val & 4319 & 500 \\
    & VQA-V2 & val & 214354 & 500 \\
    & TextVQA   & val      & 5000  & 500  \\ \midrule

\multirow{2}{*}{Math \& Science} & MathVista & testmini & 1000  & 1000 \\
            & AI2D       & test  & 3088  & 500  \\ \midrule

Visual Dialogue & LLaVA-W    & test  & 60    & 60   \\ \midrule

\multirow{6}{*}{Multi-discipline}                                          & MM-Bench       & cn-dev   & 4329  & 500 \\
            & MM-Bench       & en-dev   & 4377   & 500  \\
            & MME       & cog. \& percep.   & 2374  & 2374 \\
            & MMMU       & val   & 900   & 900  \\
            & CMMMU      & val   & 900   & 900  \\
            & Seed-Bench & test  & 17990 & 500  \\

\midrule
\cellcolor{gray!25} - & \textbf{Total} & \cellcolor{gray!25} - & \textbf{340226} & \textbf{13734} \\ \bottomrule
\end{tabular}%
}   
\end{table*}


\begin{figure*}[h] 
    \centering
    \makebox[0.3\textwidth]{\scriptsize (a) AI2D}
    \makebox[0.3\textwidth]{\scriptsize (b) Flickr30k}
    \makebox[0.3\textwidth]{\scriptsize (c) InfoVQA}
    \\
    \includegraphics[width=0.3\textwidth]{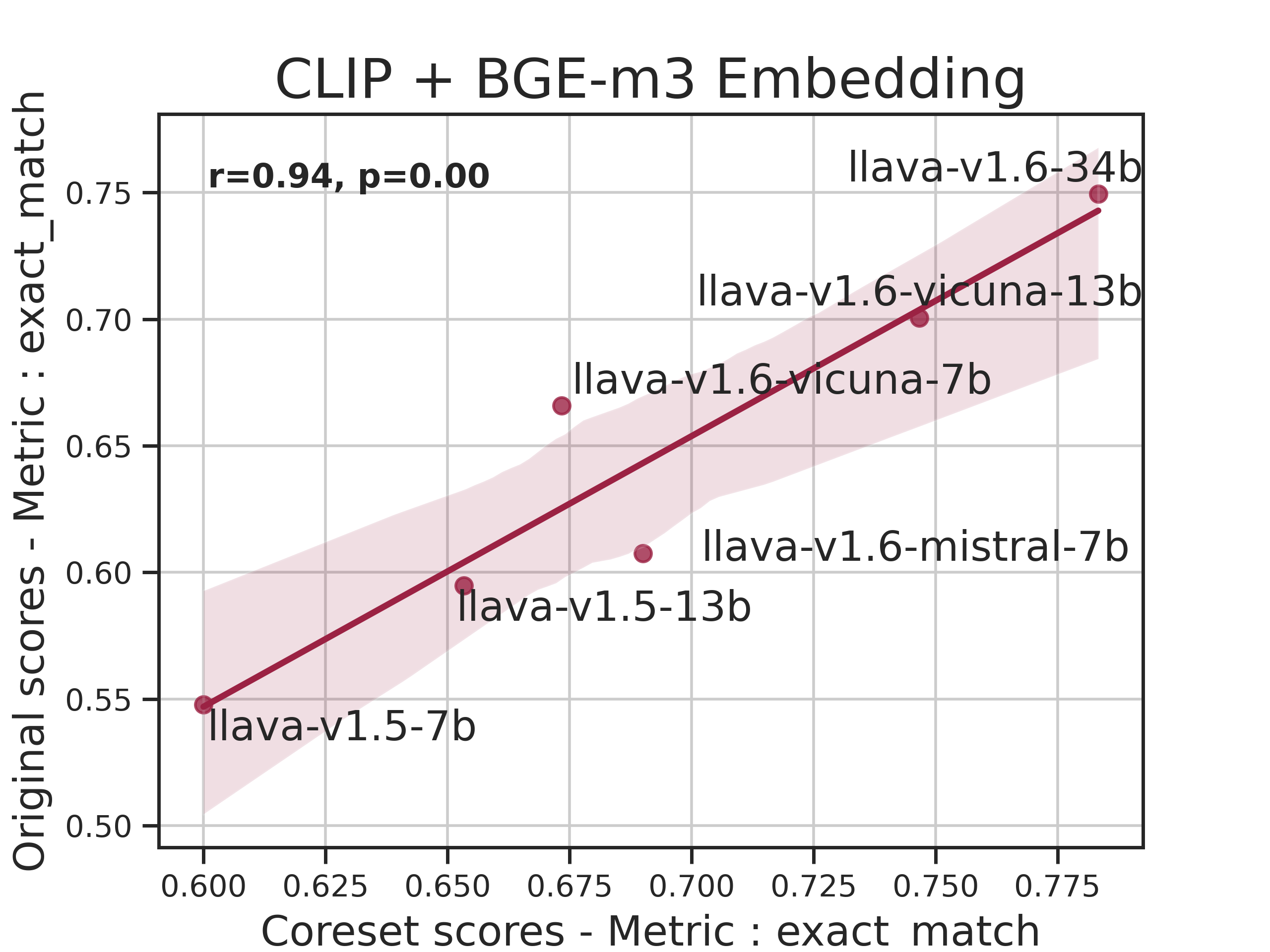}
    \includegraphics[width=0.3\textwidth]{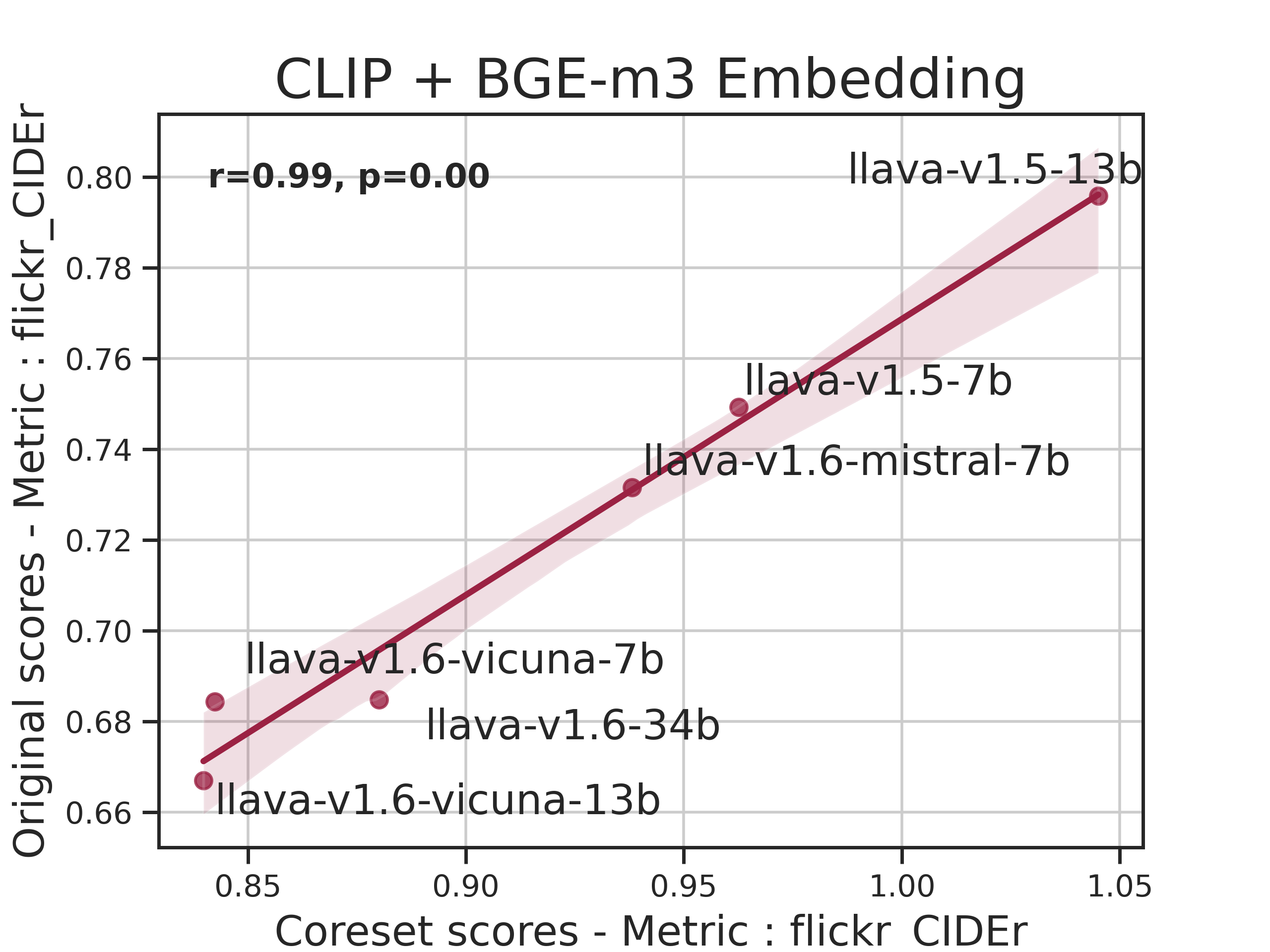}
    \includegraphics[width=0.3\textwidth]{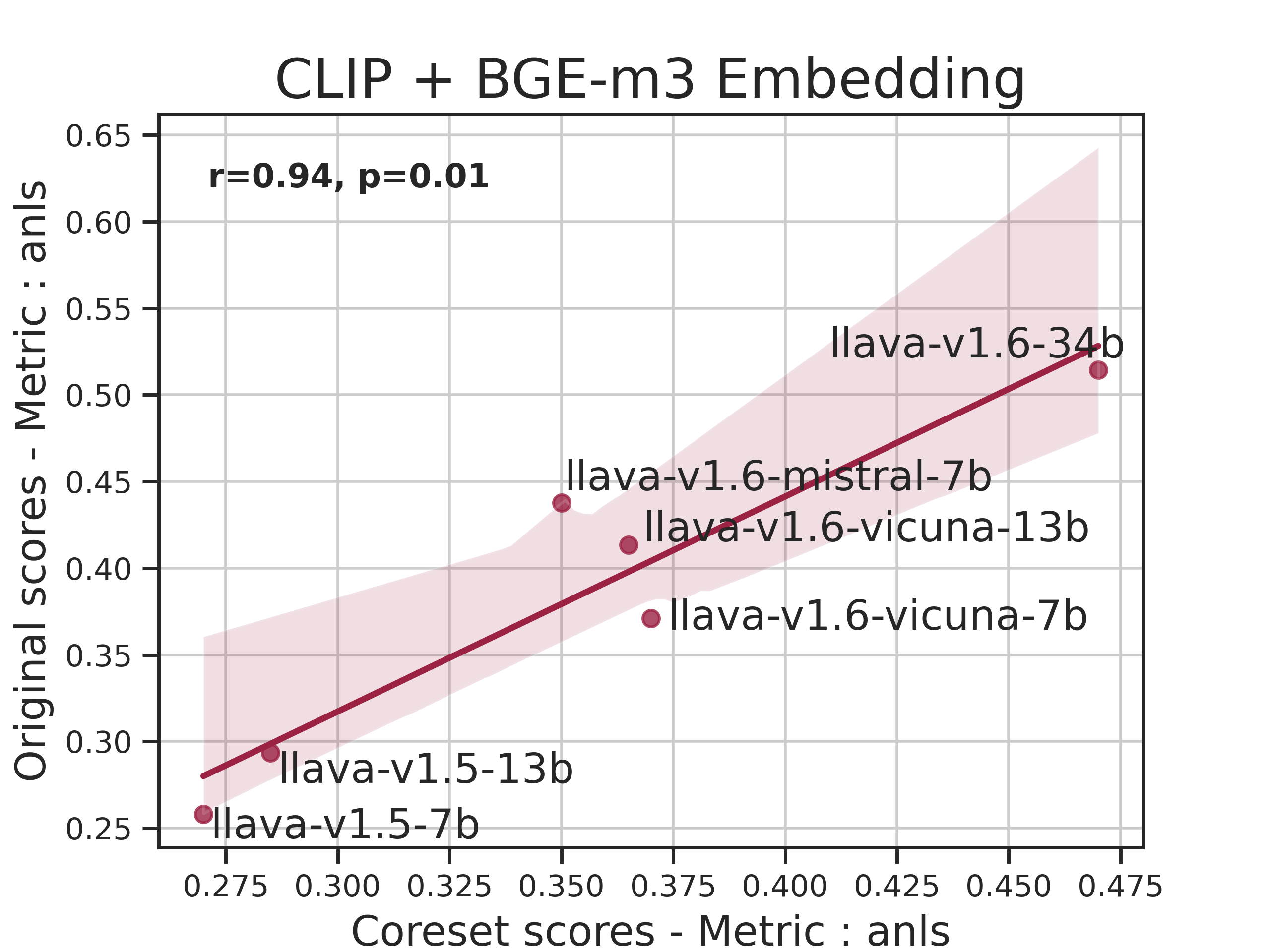}
    \\
    \caption{Correlation Graph between scores for our lite set and original scores} 
    \label{fig:correlation-graph}
\end{figure*}

We curated the first version of \lmmsevallite and present its correlation score and aggregation score in the paper. The exact plot of the correlation can be refered to \Cref{fig:correlation-graph}

\subsection{Curating more datasets in \lmmsevallite~}
\label{appendix:lmms-lite-plus}


We applied the same algorithm to additional datasets to develop a more comprehensive and diverse Lite version. In contrast to the original \lmmsevallite~, our version incorporates more datasets, including COCO \citep{lin2014microsoft} and VQA \citep{balanced_vqa_v2}.

\subsection{k-Center Greedy algorithm}
\label{appendix:greedy}

The greedy algorithm we use for $k$-center clustering is detailed in \Cref{alg:greedy}. In $k$-center clustering, the objective is to select $k$ points among $V$ vertices such that the maximum distance from any point in $V$ to its nearest cluster center is minimized. In the employed greedy algorithm, a random point is initially chosen as a center. Subsequently, the distance from this center to every other point is updated. The point with the maximum distance from the current centers is then selected and added to the center list. This process is repeated until $k$ center points have been identified.

\begin{minipage}{0.45\textwidth}
\vspace{-2mm}
   \begin{algorithm}[H] 
   \caption{$k$-Center-Greedy} 
   \label{alg:greedy} 
   \begin{algorithmic}
        \State {\bfseries Input:} data $\mathbf{x}_i$ and $\left | V \right | = n$
        \State Initialize $\mathbf{s}=\mathbf{\phi}$
        \While{$\left | s \right | < n$}
            \State $u=\arg\max_{i \in D \setminus \mathbf{s}} \min_{j \in \mathbf{s}} \Delta(\mathbf{x}_i, \mathbf{x}_j)$ 
            \State $\mathbf{s} = \mathbf{s} \cup \{u\}$ 
        \EndWhile
        \State {\bfseries return} $\mathbf{s}$
\end{algorithmic}
\end{algorithm} 
\vspace{-3mm}
\end{minipage} 

\section{LiveBench Details}

\subsection{Website Candidates}
\label{appendix:website_list}

To evaluate the performance and reliability of various news and information sources, a diverse set of websites has been selected for \livebench. We present the websites in~\Cref{tab:websites}. These websites span multiple categories, ensuring comprehensive coverage of different domains such as general news, business, technology, and international affairs. The list of candidate websites for \livebench~includes prominent sources like BBC, CNN, Bloomberg, WSJ, and Reuters, among others. Each of these websites has been categorized based on its primary content focus. This categorization aids in the systematic evaluation of the content quality and the impact of imagery and reporting styles across different domains. It should be noted that this is a initial set of candidate websites and there may be changes depending on the situations of these websites.

\subsection{Dataset Curation Prompts}
\label{appendix:curation_prompts}

This section outlines the dataset curation process, especially prompts used in different stages. First, the quiz model is provided with prompts to generate questions from raw website screenshots. The details of this prompt can be accessed at ~\cref{tab:qa-generator}.

Once the candidate QAs are generated, we instruct the models to create corresponding scoring criteria for each question. The prompt used for this process is available at ~\cref{tab:scorer}. Each question is graded on a 10-point scale based on the provided criteria.

Lastly, we employ a checking model to verify the accuracy of the generated QAs. The prompt for this step is available at this ~\cref{tab:checker}. 


\subsection{Evaluation Prompts}
\label{appendix:evalution_prompts}

We utilize GPT-4o as the default judge model due to its popularity and high-throughput API. Additionally, Claude-3.5-Sonnet and Gemini 1.5 Pro serve as alternative judge models. The final report results are scaled to an accuracy metric ranging from 0 to 100 based on the assigned scores.

Criteria are specified for each question, and we instruct the judge model to follow these criteria when determining the final score. An example of the criteria is provided at~\cref{tab:question_criteria}. Detailed judge prompts are available at ~\cref{tab:judge_prompt}.

\begin{table*}[htp]
\begin{minipage}{0.99\textwidth}
\begin{AIbox}{Critieria Example}
\centering
\scalebox{0.95}{
\begin{tabular}{l p{13.5cm}}
&  \includegraphics[width=11.5cm]{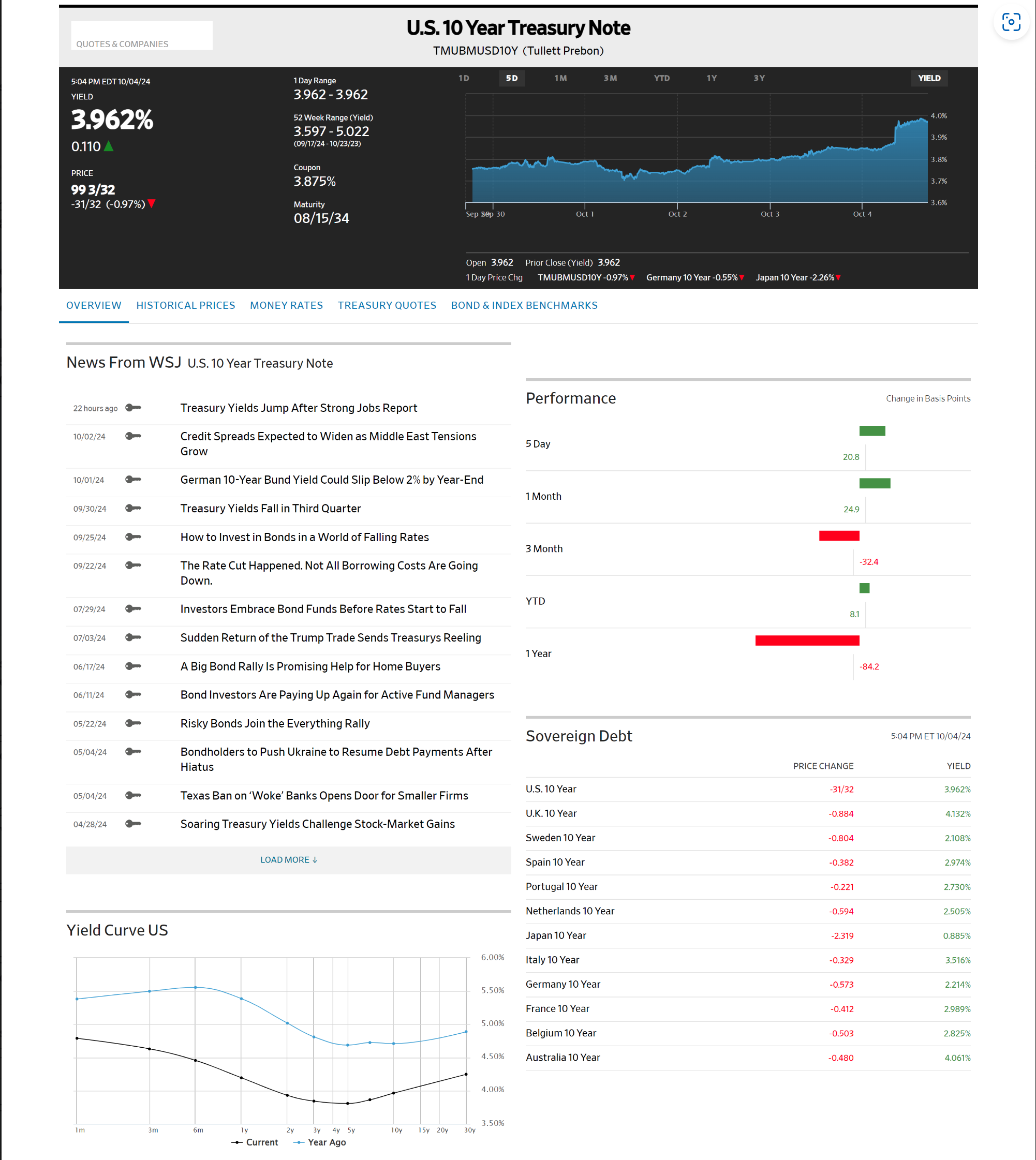} \\
\footnotesize
Question & Identify the key details of the U.S. 10 Year Treasury Note provided in the image for October 4, 2024. \\
\midrule
\footnotesize
Ground Truth & The key details of the U.S. 10 Year Treasury Note on October 4, 2024, include: - Yield: 3.962\% - Price: 99 3/32, change of -31/32 (-0.97\%) - Day Range (Yield): 3.962 - 3.962 - 52 Week Range (Yield): 3.597 - 5.022 - Coupon: 3.875\% - Maturity Date: 08/15/34 \\
\midrule
\footnotesize
Criteria & The scoring criteria for the answer are as follows: 1. Yield: 2 points - 1 point for the correct yield value - 1 point for the correct formatting 2. Change in Yield: 2 points - 1 point for the correct change value - 1 point for the correct formatting 3. Price: 2 points - 1 point for the correct price value - 1 point for the correct formatting with the change 4. Day Range (Yield): 1 point - 1 point for the correct values 5. 52 Week Range (Yield): 1 point - 1 point for the correct values 6. Coupon: 1 point - 1 point for the correct value 7. Maturity Date: 1 point - 1 point for the correct value and formatting Deductions: - Minor mistakes in values: Deduct 0.5 points for up to 2 mistakes - Minor mistakes in formatting: Deduct 0.5 points for up to 2 mistakes \\

\end{tabular}
}
\end{AIbox}
\captionof{table}{An example of question, answer and criteria in LiveBench-09}
\vspace{-6mm}
\label{tab:question_criteria}
\end{minipage}
\end{table*}


\subsection{Question Categorization and Examples}
\label{appendix:livebench_examples}

Building upon the principles of Bloom's Taxonomy~\citep{bloom1956taxonomy}, we aim to investigate the types of information that readers can extract from news content at different cognitive levels. Specifically, we focus on how readers interpret and process news reports, categorizing the information into the following hierarchical levels:

\textbf{Concrete Recognition}: At this level, the goal is to recognize facts and explain the fundamental concepts conveyed in the news. This may require models to possess optical character recognition (OCR) capabilities to comprehend the context from provided screenshots and conclude the information. Example questions include: \textit{What are the key points in this news story?} and \textit{How would you explain the main event reported here?}

\textbf{Realworld Application}: At this level, individuals apply knowledge to real-world situations. Example questions include: \textit{Please present this news in Arabic and output it in markdown format}, \textit{Organize all the news on this page in the form of an HTML table, including the title, release time, and keywords}, \textit{Sort out the exchange rate data and plot them using the Julia language}, \textit{Please write a summary of the news in Vietnamese}, and \textit{Can you give me an example of this update in Python?}

\textbf{Analytical Understanding}: This intermediate level emphasizes dissecting the news content to understand relationships and deeper meanings. Questions at this stage encourage analysis of the factors leading to an event and how it connects with other current issues. Example questions include: \textit{What are the factors that led to this event?} and \textit{How does this event relate to other current issues?}

\textbf{Divergent Thinking \& Creation}: At the highest level, individuals engage in generating new ideas and synthesizing concepts to produce creative solutions. Questions at this level are designed to inspire divergent thinking and originality. Example questions include: \textit{How could you create a new headline that captures the essence of the event differently?} and \textit{If you were the reporter, how would you approach this story to provide a unique angle?}

We evaluate the model's performance across these four progressively challenging levels, allowing us to assess its ability to transition from basic understanding to higher-order reasoning and creative thinking.

Specific examples corresponding to these levels are provided below. Tables~\ref{tab:question_understanding}, \ref{tab:question_application}, \ref{tab:question_analytical}, and \ref{tab:question_creation} present representative examples within the \texttt{LiveBench-2024-09} evaluation, illustrating the spectrum of cognitive demands posed by each level.

\subsection{Case Analysis on \livebench}
\label{subsec:livebench-case-analysis}

We present failure case analyses in ~\cref{tab:case-LLaVA-OV,tab:case-LLaMA-vision,tab:case-LLaVA-1.5,tab:case-LLaVA-OV-Ana} to illustrate instances where current LMMs fail to respond accurately in our benchmark and the gap between these models and GPT-4o. 

In ~\cref{tab:case-LLaVA-1.5}, the model can not understand Japanese correctly and thus producing repeated nonsense sentences.

In ~\cref{tab:case-LLaVA-OV-Ana}, we see that the open-source model firstly made errors on identifying the correct numbers of the closing prices and then can not perform the arithmetic operations to get the average price, while GPT-4o manages to do so.

In ~\cref{tab:case-LLaVA-OV}, the model incorrectly matched the player names and their opponents. For instance, Karolina Muchova was supposed to play against Qinwen Zheng, but the model incorrectly stated that Muchova was leading against Anna Blinkova. Additionally, the model misidentified Qinwen Zheng as Qiang Wang, another Chinese tennis player. This demonstrates the model's difficulty in recognizing small text on websites and its tendency to hallucinate when failing to understand the image.

In ~\cref{tab:case-LLaMA-vision}, the model provided a detailed description but failed to summarize the main points. This indicates that the model may sometimes output unnecessary information and struggles with following instructions effectively.

\begin{table*}
\begin{minipage}{0.99\textwidth}
\begin{AIbox}{Prompt : QA Generator}
\centering
\begin{lstlisting}[basicstyle=\tiny,]
I would like you to act as a quizmaster who designs questions based on a provided image that would challenge adults to think critically. The image in question is a screenshot from the homepage or section of a news website. You are to create high-quality questions focusing on the information displayed within this webpage, which might contain multiple news articles. Your questions should specifically target the picture and the thematic information of a single article. Your question should be answerable, and checkable. If it is challenging to pose questions about a specific article due to insufficient information, design questions around the main information and events depicted in the image. Within your capabilities, try to make the questions more challenging.

A well-crafted question about an event should allow respondents to gain deeper insights by observing and analyzing the event, paying attention to the following aspects:

1. **Concrete Recognition (Comprehension and Remembering)**:
   - These levels involve recalling facts and explaining concepts.
   - Example questions include:
     - "What are the key points in this news story?" (Remembering)
     - "How would you explain the main event reported here?" (Comprehension)
2. **Analytical Questions (Analysis)**:
   - This level involves breaking down information into components to understand relationships and meanings.
   - Example questions:
     - "What are the factors that led to this event?"
     - "How does this event relate to other current issues?" 
3. **Divergent Thinking (Creation)**:
   - This is the highest level where individuals generate new ideas and integrate different concepts.
   - Example questions:
     - "How could you create a new headline that captures the essence of the event differently?"
     - "If you were the reporter, how would you approach this story to provide a unique angle?"
     - "Do you think the report's presentation was fair? Why or why not?"
4. **Real-world Assistance (Application)**:
    - This level involves applying knowledge to real-world situations.
    - Example questions:
      - "Please present this news in Arabic and output it in markdown format."
      - "Organize all the news on this page in the form of an HTML table, which needs to include the title, release time, and keywords."
      - "Sort out the exchange rate data and plot them using Julia language."
      - "Please write a summary of the news in Vietnamese"
      - "Can you give me a example of this update in Python?" (Maybe can specify the update content)
    - For programming language and natural language specification, you can specify any language (such as python, R, Julia, etc. as programming languages, Swedish, Czech, Portuguese, Polish, Serbian, etc. as natural languages).

Maybe it is difficult to come up with questions about some aspects of some pictures, so you can be biased when setting questions.

Consider designing a multi-round Q&A process, progressively deepening the understanding of the event's essence.

Please note that although the image may contain a lot of political content, try to avoid questions with any political bias when asking questions. Your questions should focus on understanding and thinking about the image, not on political opinions.

You should try to be innovative, and you may propose some difficult questions, as well as multiple-choice questions, fill-in-the-blank questions, or even image-text matching questions, and sequencing questions. Within your capabilities, try to make the questions more challenging.

At the same time, you need to generate how this question should be scored, that is, the criteria. Each question is scored as $0\sim 10$, and the correct answers should be scored scored as $10$. Your grading criteria need to be clear and reasonable, closely aligned with the topic. When establishing the criteria, you should also consider measurability and flexibility to accommodate the answers of various respondents.
\end{lstlisting}
\end{AIbox}
\end{minipage}
\caption{The prompt that use to generate QA pairs}
\label{tab:qa-generator}
\end{table*}

\begin{table*}
\begin{minipage}{0.99\textwidth}
\begin{AIbox}{Prompt : Scorer}
\centering
\begin{lstlisting}[basicstyle=\tiny,]
Based on the multi-round Q&A regarding the image, please evaluate each question and answer from the multi-round Q&A based on the image for their authenticity (whether the information can be directly obtained from the image or reasonably inferred) and logical coherence. For each Q&A pair, provide a rating from 1 to 10, where 1 indicates very poor and 10 indicates excellent. Additionally, please provide a brief explanation for each rating.

Here are the criteria for evaluating the Q&A pairs:

### 1. Authenticity (5 points)
- **5 Points**: The information is directly observable in the image or can be reasonably inferred with strong evidence.
- **3 Points**: The information has a plausible connection to the image but requires assumptions that are not strongly supported by the image.
- **1 Point**: The information cannot be observed or reasonably inferred from the image; it seems unrelated or speculative.

### 2. Logical Coherence (3 points)
- **3 Points**: The answer logically follows from the question and maintains consistency with the image context.
- **2 Points**: There are minor logical gaps or inconsistencies in the answer relative to the question.
- **1 Point**: The answer is logically inconsistent or contradictory to the question or image context.

### 3. Clarity and Precision (2 points)
- **2 Points**: The question and answer are clearly articulated and precisely address specifics of the image.
- **1 Point**: The question or answer is somewhat vague or overly general, lacking specific details related to the image.
- **0 Points**: The question or answer is unclear or too ambiguous to determine relevance to the image.

Each Q&A pair can score a maximum of 10 points. The sum of points from these three categories determines the final score for each pair. Provide a brief explanation for each rating, focusing on how well the Q&A adheres to these criteria.
\end{lstlisting}
\end{AIbox}
\end{minipage}
\caption{The prompt that use to score the QA pairs}
\label{tab:scorer}
\end{table*}

\begin{table*}
\begin{minipage}{0.99\textwidth}
\begin{AIbox}{Prompt : Checker}
\centering
\begin{lstlisting}[basicstyle=\tiny,]
I would like you to act as a quizmaster who designs questions based on a provided image that would challenge adults to think critically. The image in question is a screenshot from the homepage or section of news website. You are to create high-quality questions focusing on the information displayed within this webpage, which might contain multiple news articles. Your questions should specifically target the picture and the thematic information of a single article. Your question should be answerable, and checkable. Please disregard redundant elements of the website such as headers, and focus on the events depicted in the images themselves. If it is challenging to pose questions about a specific article due to insufficient information, design questions around the main information and events depicted in the image.

Now, you are given a screenshot of the homepage of a news website, with a already generated question and answer. Your task is to refine the question and answer, and refractor them to make the question more answerable, checkable, and challenging. If you don't think the question is good, please provide a new question and answer.

Note that the subtask must be one of these four:

- Concrete Recognition
- Analytical Questions
- Divergent Thinking
- Real-world Assistance

If you think the question does not correspond to the subtask, you have two options:
1. Modify the question to correspond to the subtask.
2. Modify the subtask to correspond to the question.

However, you should not change the original question's subtask unless the original subtask is not one of these five. If you feel the original question's subtask does not match the question, modify the question to match the subtask instead of rewriting the subtask.

Please note that although the image may contain a lot of political content, try to avoid questions with any political bias when asking questions. The question should focus on understanding and thinking about the image, not on political opinions. Within your capabilities, try to make the questions more challenging. However, you also need to consider the gradability of the questions you set. It is reiterated that what you need to assess is the ability to understand the news webpage, not politics.

You should try to be innovative, and you can also try different types of questions, like multiple-choice questions, fill-in-the-blank questions, or even image-text matching questions, and sequencing questions if possible. Within your capabilities, try to make the questions more challenging.

If you think the question is not good, or it is not answerable, please provide a new question and answer.

Reminder again that you cannot change the original subtask unless the original subtask is not one of the five listed above.
\end{lstlisting}
\end{AIbox}
\end{minipage}
\caption{The prompt that use to check the QA pairs}
\label{tab:checker}
\end{table*}

\begin{table*}
\begin{minipage}{0.99\textwidth}
\begin{AIbox}{Prompt : Finalizer}
\centering
\begin{lstlisting}[basicstyle=\tiny,]
You are a question setter, and your task is to finalize the question, answer, and scoring criteria. Make sure:

1. The criteria should be a natural language, don't use dict / json format for the criteria, human cannot understand it.
2. You can use bullet points / numbers to the list / yaml format to the criteria. But don't use python-like format.
3. If the answer is in dict format, but there is no need to answer in dict format (means there is a way to answer in natural language, the question do not specify to answer in dict format), you should convert it to natural language.
4. If the whole criteria is in other language, change it to English. But if you think some words should be in other language, you can keep it in that language. If question or answer is in other language, you don't need to change it.
5. The scoring criteria are rational and facilitate the accurate assessment of responses.
6. The full score for the scoring criteria must be 10 points, and it must directly relate to the specific answer.
7. The question is clear and unambiguous.
8. The answer is correct and reasonable (although the original ground truth answer is mostly correct, it may not be perfect, and sometimes the answer maybe incorrect).

Some tips:

1. For some extremely hard open-ended questions where answers may vary, hitting all points perfectly may not be realistic. In such cases, you can relax the criteria slightly. For example, if there are five possible points in an answer, but answering three adequately could merit full points. An other option is to change the question to a multiple-choice / multi-select question. But remember, it only applies to extremely hard open-ended questions which are impossible to answer perfectly.
2. For some questions, changing the format might be beneficial. You can consider transforming them into different types of questions such as essay, fill-in-the-blank, ranking (e.g., based on time, importance, etc.), or matching questions to enhance the difficulty and rationality of the scoring criteria. But a very important point is that DO NOT CHANGE the question to multiple-choice questions. If the original question is multiple-choice, you need to change it to another type of question (e.g., open-source, fill-in-the-blank, etc.).
\end{lstlisting}
\end{AIbox}
\end{minipage}
\caption{The prompt that use to finalize the QA pairs}
\label{tab:finalizer}
\end{table*}

\begin{table*}
\begin{minipage}{0.99\textwidth}
\begin{AIbox}{Prompt : Information Extractor}
\centering
\begin{lstlisting}[basicstyle=\tiny,]
These are the images of the website that we have captured. Please extract the text from the website.
You should extract the text from the website as detailed as possible.
Only output the text extracted from the website, do not include any other information.
\end{lstlisting}
\begin{lstlisting}[basicstyle=\tiny,]
This is a screenshot from a news website. Your task is to identify the meaningful images in this screenshot and extract relevant information about these images, such as the environment depicted, the actions and expressions of the people, and the connection between these images and the corresponding text. You need to think deeply about these images and provide as much detailed and useful information as possible. Of course, it is also possible that the website is mainly text-based, and in this case, there might not be much information to extract from the images. In such instances, you can approach the task from a textual perspective, analyzing the website's content. For example, what is the theme of the website? What kind of information is contained in the text? If the website requires thoughtful analysis, feel free to engage in deeper reflection and provide your insights.
\end{lstlisting}
\begin{lstlisting}[basicstyle=\tiny,]
What makes this website different from other websites? What is special about its news? Since it is a news website, where is the 'new' aspect reflected? Do not provide a generalized answer; you need to give detailed responses based on the specific content of each news article and the accompanying illustrations.

For example, if the news is about a software update, what conveniences will this update bring to people? How can people use these new features? Perhaps there are also some drawbacks? You need to come up with your own questions worth pondering about the website and describe in as much detail as possible your understanding of what is 'new' on the website.
\end{lstlisting}
\end{AIbox}
\end{minipage}
\caption{The prompt that use to extract information from website}
\label{tab:info-extractor}
\end{table*}

\begin{table*}
\begin{minipage}{0.99\textwidth}
\begin{AIbox}{Prompt : Information Extractor}
\centering
\begin{lstlisting}[basicstyle=\tiny,]
[Question]

{prompt}

[Assistant Response]

{generation}

[Ground Truth Response]

{reference}

[System]

Rate whether the assistant response correctly matches the ground truth, in regards to the image above.

The rating should be 0-10, where 0 is incorrect and 10 is correct.

Below is the specific criteria for rating:

{criteria}

Your response should be in the JSON format:
```json
{{
    "Explanation": "(your explanation)",
    "Rating": "(int)"
}}
```
\end{lstlisting}
\end{AIbox}
\end{minipage}
\caption{The judge prompt that used in evaluation.}
\label{tab:judge_prompt}
\end{table*}

\begin{table*}[htp]
\begin{minipage}{0.99\textwidth}
\begin{AIbox}{Case Analysis : LLaVA-1.5-7B (Concrete Recognition) }
\centering
\scalebox{0.80}{
\begin{tabular}{l p{13.5cm}}
&  \includegraphics[width=13.5cm]{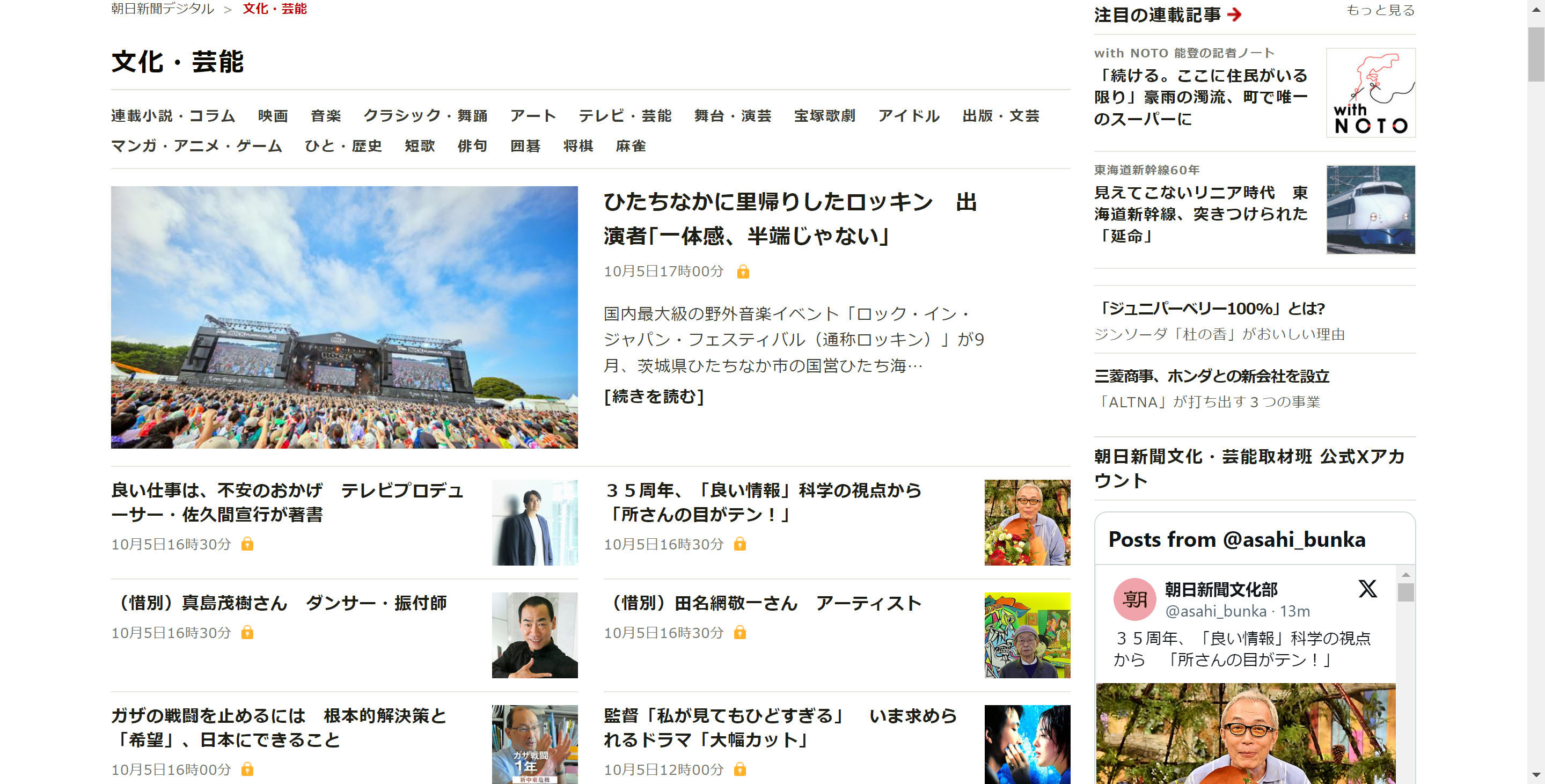} \\
\footnotesize
Question & Sort all the news headlines that appear in the image from newest to oldest. Provide the headlines and their respective timestamps. \\
\midrule
\footnotesize
LLaVA-1.5-7B & 1. 2014/09/26 14:00:00 - 2014/09/27 00:00:00 2. 2014/09/26 13:00:00 - 2014/09/27 00:00:00 3. 2014/09/26 12:00:00 - 2014/09/27 00:00:00 4. 2014/09/26 11:00:00 - 2014/09/27 00:00:00 5. 2014/09/26 10:00:00 - 2014/09/27 00:00:00 6. 2014/09/26 09:00:00 - 2014/09/27 00:00:00 7. 2014/09/26 08:00:00 - 2014/09/27 00:00:00 8. 2014/09/26 07:00:00 - 2014/09/27 00:00:00 9. 2014/09/26 06:00:00 - 2014/09/27 00:00:00 10. 2014/09/26 05:00:00 - 2014/09/27 00:00:00 11. 2014/09/26 04:00:00 - 2014/09/27 00:00:00 12. 2014/09/26 03:00:00 - 2014/09/27 00:00:00 13. 2014/09/26 02:00:00 - 2014/09/27 00:00:00 14. 2014/09/26 01:00:00 - 2014/09/27 00:00:00 15. 2014/09/26 00:00:00 - 2014/09/27 00:00:00 \\
GPT4o-Score & 0 \\
GPT4o-Rationale & The assistant response did not match the ground truth at all. The assistant provided a list of timestamps that do not correspond to the actual timestamps in the image. Additionally, the headlines provided by the assistant are completely incorrect and do not match any of the headlines in the image. Therefore, the response is entirely incorrect. \\
GPT4o-Response & \begin{CJK}{UTF8}{goth}Here are the news headlines sorted from newest to oldest along with their respective timestamps: 1. ひたちなかに里帰りしたロッキン·出演者「一体感、半端じゃない」 - October 5, 17:00 2. 良い仕事は、不安のおかげ·テレビプロデューサー・佐久間宣行が著書 - October 5, 16:30 3. ３５周年、「良い情報」科学の視点から「所さんの目がテン！」 - October 5, 16:30 4. (惜別) 真島茂樹さん·ダンサー・振付師 - October 5, 16:00 5. ガザの戦闘を止めるには·根本的解決策と「希望」、日本にできること - October 5, 16:00 6. (惜別) 田名網敬一さん·アーティスト - October 5, 16:00 7. 監督「私が見てもひどすぎる」·いま求められるドラマ「大幅カット」 - October 5, 12:00 \end{CJK} \\
\end{tabular}
}
\end{AIbox}
\captionof{table}{An example of the failure case of LLaVA-1.5-7B in LiveBench-09}
\vspace{-6mm}
\label{tab:case-LLaVA-1.5}
\end{minipage}
\end{table*}

\begin{table*}[htp]
\begin{minipage}{0.99\textwidth}
\begin{AIbox}{Case Analysis : Qwen-VL-72B-Instruct (Analytical Questions) }
\centering
\scalebox{0.80}{
\begin{tabular}{l p{13.5cm}}
&  \includegraphics[width=13.5cm]{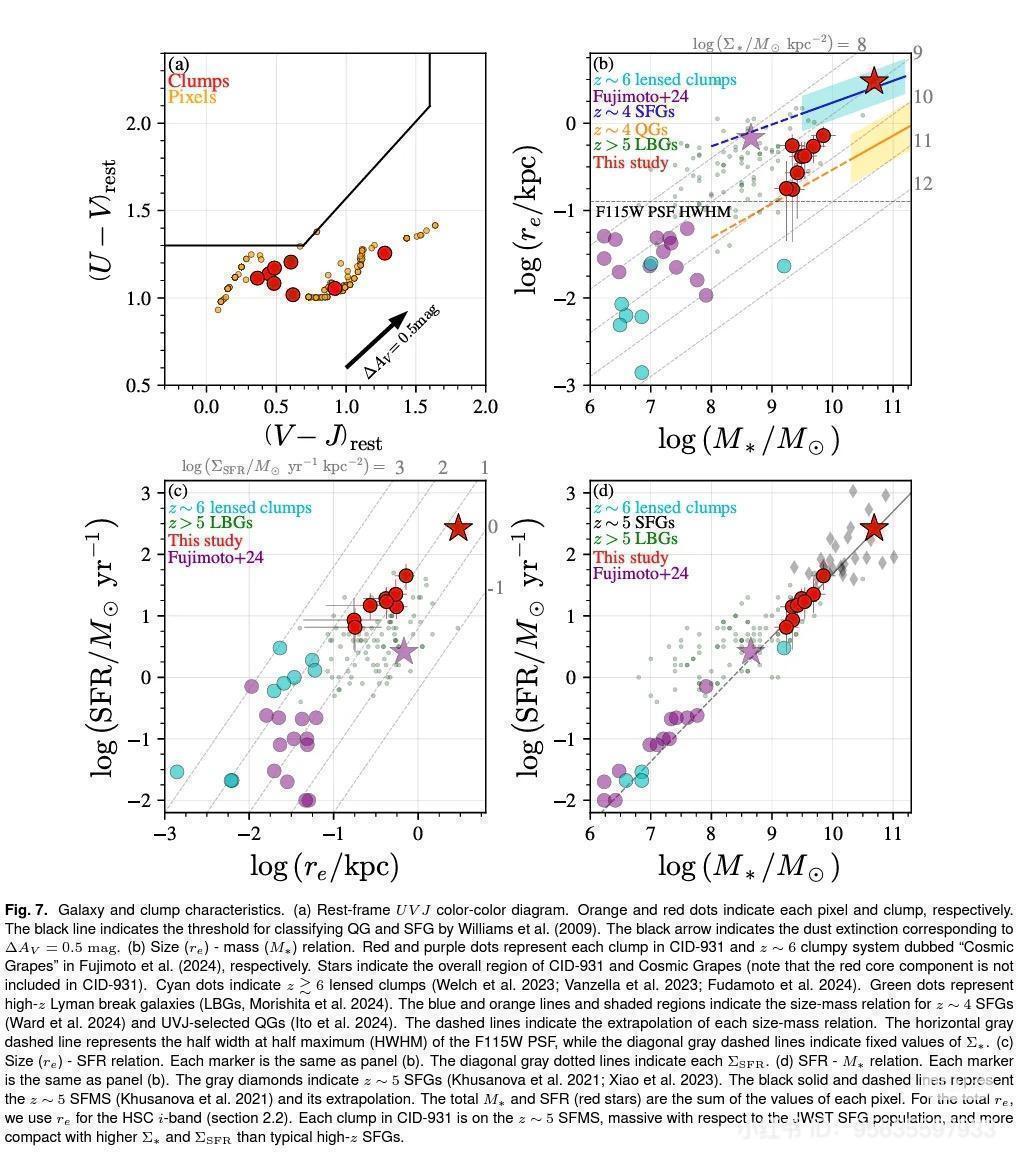} \\
\footnotesize
Question & Based on graph (b) in the provided figure, what are the key differences between the galaxy characteristics of the 'cosmic grapes' clumps and the 'CID-931 clumps'? \\
\midrule
\footnotesize
Qwen-VL-72B-Instruct & Graph (b) shows the size ($r_e$) - mass ($M_*$) relation for different galaxy populations. The 'cosmic grapes' clumps are represented by red dots, while the 'CID-931 clumps' are represented by purple dots. The key differences between the two populations are: 1. The 'cosmic grapes' clumps have a larger range of sizes ($r_e$) compared to the 'CID-931 clumps'. 2. The 'cosmic grapes' clumps have a larger range of masses ($M_*$) compared to the 'CID-931 clumps'. 3. The 'cosmic grapes' clumps are generally larger and more massive than the 'CID-931 clumps'.\\
GPT4o-Score & 1 \\
GPT4o-Rationale & The assistant's response does not correctly match the ground truth. The assistant incorrectly states that the 'cosmic grapes' clumps have a larger range of sizes and masses compared to the 'CID-931 clumps'. Additionally, the assistant does not address the surface density differences between the two populations. The ground truth correctly identifies that 'CID-931 clumps' have higher stellar masses, larger sizes, and higher surface densities compared to 'cosmic grapes'. \\
\end{tabular}
}
\end{AIbox}
\captionof{table}{An example of the failure case of Qwen-VL-72B-Instruct in LiveBench-09}
\vspace{-6mm}
\label{tab:case-Qwen-72B-AnaQues}
\end{minipage}
\end{table*}

\begin{table*}[htp]
\begin{minipage}{0.99\textwidth}
\begin{AIbox}{Case Analysis : Qwen-VL-72B-Instruct (Realworld Assistance) }
\centering
\scalebox{0.80}{
\begin{tabular}{l p{13.5cm}}
&  \includegraphics[width=13.5cm]{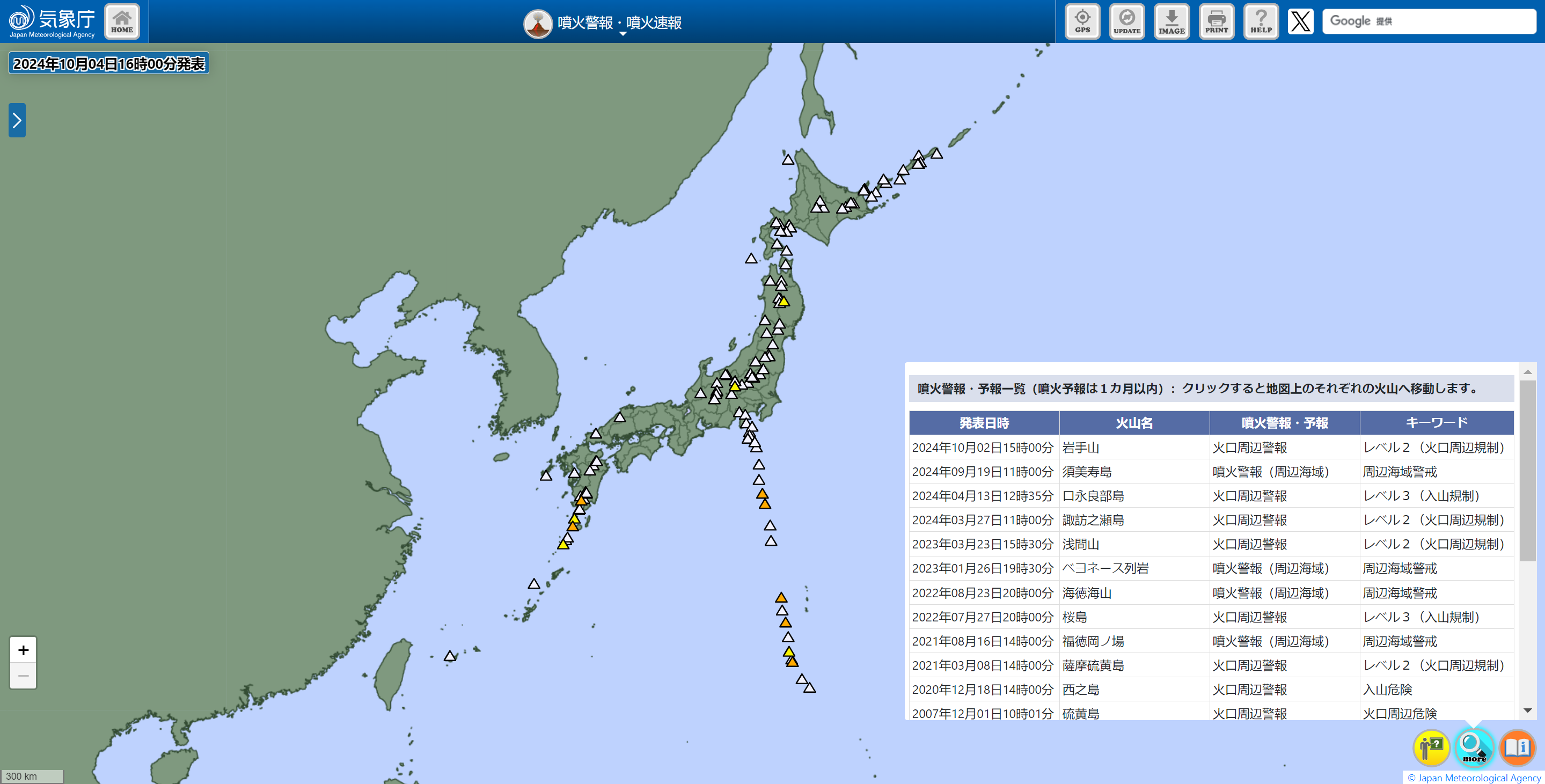} \\
\footnotesize
Question & Based on the volcanic warnings listed on the Japan Meteorological Agency's website, identify the volcano that was issued a Level 3 warning on April 13, 2024. Additionally, describe the type of warning and any precautions that were mentioned. \\
\midrule
\footnotesize
Qwen-VL-72B-Instruct & 	
\begin{CJK}{UTF8}{goth} The volcano that was issued a Level 3 warning on April 13, 2024, is 水良部島 (Mizunoshima Island). The type of warning is "火山周辺警報" (Volcanic Surrounding Alert), and the precautions mentioned are "レベル 3（入山規制）" (Level 3 (Mountain Entry Restriction)). \end{CJK}\\
GPT4o-Score & 0 \\
GPT4o-Rationale & \begin{CJK}{UTF8}{goth}The assistant incorrectly identified the volcano as 水良部島 (Mizunoshima Island) instead of 口永良部島 (Kuchinoerabujima), which results in 0 points for the first criterion. The type of warning was also incorrectly described as '火山周辺警報' (Volcanic Surrounding Alert) instead of '噴火周辺警報' (Eruption Warning - Surrounding Areas), resulting in 0 points for the second criterion. The assistant did not mention any of the specific precautionary measures (evacuation, restriction on entry, vigilance), resulting in 0 points for the third criterion.\end{CJK} \\
\end{tabular}
}
\end{AIbox}
\captionof{table}{An example of the failure case of Qwen-VL-72B-Instruct in LiveBench-09}
\vspace{-6mm}
\label{tab:case-Qwen-72B-RealworldAsst}
\end{minipage}
\end{table*}

\begin{table*}[htp]
\begin{minipage}{0.99\textwidth}
\begin{AIbox}{Case Analysis : LLaVA-NeXT-OV-72B-Chat (Concrete Recognition) }
\centering
\scalebox{0.80}{
\begin{tabular}{l p{13.5cm}}
&  \includegraphics[width=13.5cm]{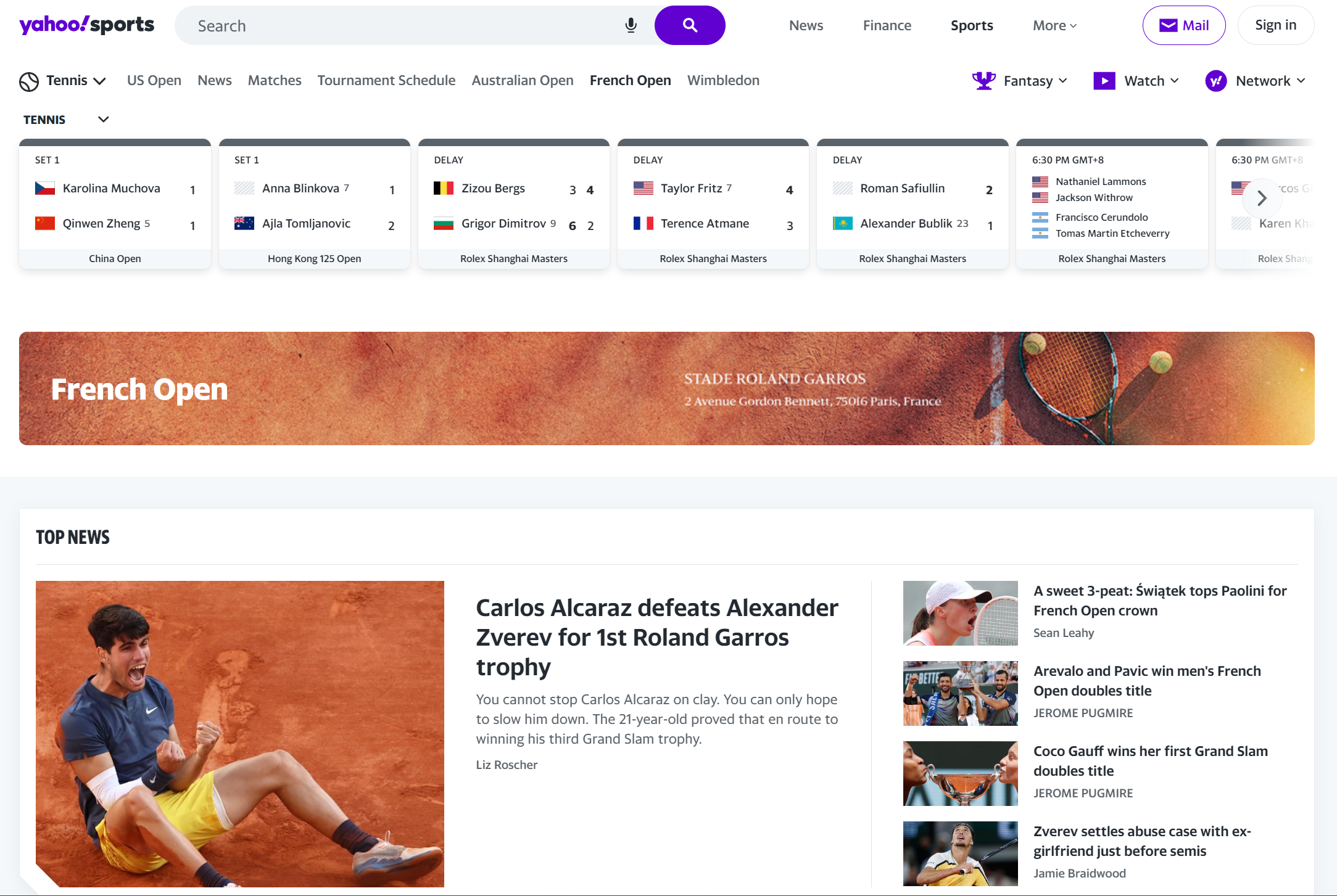} \\
\footnotesize
Question & Analyze the ongoing tennis matches displayed on the webpage, detailing the players involved, their current scores, and the tournaments they are part of. \\
\midrule
\footnotesize
LLaVA-NeXT-OV-72B-Chat & The webpage displays several ongoing tennis matches from various tournaments. In the China Open, Karolina Muchova is leading against Anna Blinkova with a score of 1-0 in the first set. Meanwhile, in the Hong Kong 125K Open, Ajla Tomljanovic is trailing against Qiang Wang with a score of 0-1 in the first set. Over at the Rolex Shanghai Masters, Zelou Bergs is leading against Taylor Fritz with a score of 3-1 in the first set, and Grigor Dimitrov is leading against Alejandro Davidovich Fokina with a score of 2-0 in the first set. In the same tournament, Stefanos Tsitsipas is leading against Pablo Cuevas with a score of 2-0 in the first set, and Alexander Zverev is leading against Frances Tiafoe with a score of 2-0 in the first set. Lastly, in the Rolex Shanghai Masters, Rafael Nadal is leading against Jackson Withrow with a score of 2-0 in the first set, and Tomas Martincak is leading against Francisco Cerundolo with a score of 2-0 in the first set. These matches showcase a mix of established and up-and-coming players, highlighting the competitive nature of these tournaments. \\
GPT4o-Score & 0 \\
GPT4o-Rationale & The assistant response incorrectly identified the players, scores, and tournaments. The correct matches and scores are: Karolina Muchova vs. Qinwen Zheng (1-1), Anna Blinkova vs. Ajla Tomljanovic (7-2), Zizou Bergs vs. Grigor Dimitrov (3-4, delayed), Taylor Fritz vs. Terence Atmane (7-4, delayed), and Roman Safiullin vs. Alexander Bublik (2-1, delayed). The upcoming matches are Nathaniel Lammons/Jackson Withrow vs. Francisco Cerundolo/Tomas Martin Etcheverry and Marcos Giron vs. Karen Khachanov. The assistant response did not match any of these details correctly. \\
GPT4o-Response & Here are the details of the ongoing tennis matches displayed on the webpage: 1. **China Open** - **Karolina Muchova** vs. **Qinwen Zheng** - Current Score: 1-1 in Set 1 2. **Hong Kong 125 Open** - **Anna Blinkova** vs. **Ajla Tomljanovic** - Current Score: 7-2 in Set 1 3. **Rolex Shanghai Masters** - **Zizou Bergs** vs. **Grigor Dimitrov** - Current Score: 3-4 in Set 1 (Match is delayed) - **Taylor Fritz** vs. **Terence Atmane** - Current Score: 7-3 in Set 1 (Match is delayed) - **Roman Safiullin** vs. **Alexander Bublik** - Current Score: 2-1 in Set 1 (Match is delayed) Upcoming Matches: - **Nathaniel Lammons / Jackson Withrow** vs. **Francisco Cerundolo / Tomas Martin Etcheverry** - Scheduled for 6:30 PM GMT+8 at the Rolex Shanghai Masters - **Marcos Giron** vs. **Karen Khachanov** - Scheduled for 6:30 PM GMT+8 at the Rolex Shanghai Masters \\
\end{tabular}
}
\end{AIbox}
\captionof{table}{An example of the failure case of LLaVA-NeXT-OV-72B in LiveBench-09}
\vspace{-6mm}
\label{tab:case-LLaVA-OV}
\end{minipage}
\end{table*}

\begin{table*}[htp]
\begin{minipage}{0.99\textwidth}
\begin{AIbox}{Case Analysis : LLaMA-3.2-Vision-11B-Instruct (Concrete Recognition) }
\centering
\scalebox{0.80}{
\begin{tabular}{l p{13.5cm}}
&  \includegraphics[width=13.5cm]{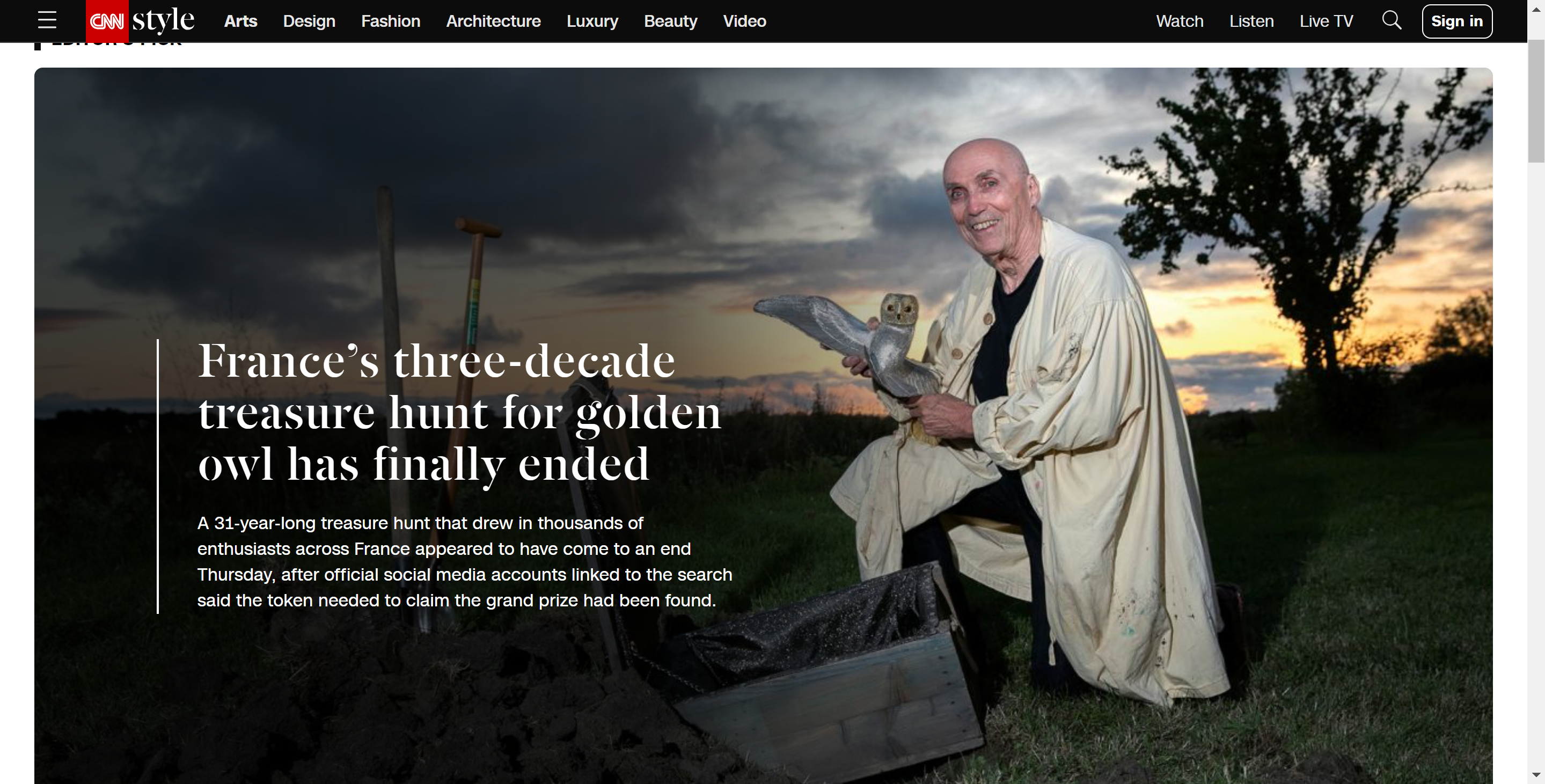} \\
\footnotesize
Question & Summarize the main points of the news article regarding the golden owl treasure hunt in France. Be sure to include the duration of the hunt, the level of participation, the conclusion, and how the conclusion was announced. \\
\midrule
\footnotesize
LLaMA-3.2-Vision-11B-Instruct & The image shows a screenshot of an article from CNN Style, featuring a man in a white coat holding a golden owl statue. The headline reads "France's three-decade treasure hunt for golden owl has finally ended." * The man is wearing a white coat and holding a golden owl statue. * He is kneeling on the grass, with his left leg bent at a 90-degree angle. * He is holding the owl statue in both hands, with his arms crossed over his chest. * The background of the image is a sunset over a field, with trees and a cloudy sky. * The sky is orange and yellow, with dark clouds in the distance. * There are trees in the background, partially obscured by the man and the owl statue. * The overall atmosphere of the image is one of excitement and celebration, as the man has finally found the treasure after a long search. * The man's smile and the golden owl statue suggest that he is happy and proud of his discovery. * The sunset in the background adds to the sense of triumph and accomplishment. The image suggests that the man has been searching for the golden owl for a long time and is finally able to hold it in his hands. The sunset in the background adds a sense of drama and excitement to the scene, emphasizing the significance of the discovery. \\
GPT4o-Score & 0 \\
GPT4o-Rationale & The assistant's response provides a detailed description of the image but fails to summarize the main points of the news article. It does not mention the duration of the hunt, the level of participation, the conclusion, or how the conclusion was announced. Therefore, it does not match the ground truth. \\
GPT4o-Response & The news article reports that France's golden owl treasure hunt, which lasted for 31 years and attracted thousands of enthusiasts across the country, has finally concluded. The end of the hunt was announced on Thursday through official social media accounts associated with the search, stating that the token required to claim the grand prize had been found. \\
\end{tabular}
}
\end{AIbox}
\captionof{table}{An example of the failure case of LLaMA-3.2-Vision-11B-Instruct in LiveBench-09}
\vspace{-6mm}
\label{tab:case-LLaMA-vision}
\end{minipage}
\end{table*}

\begin{table*}[htp]
\begin{minipage}{0.99\textwidth}
\begin{AIbox}{Case Analysis : LLaVA-NeXT-OV-72B-Chat (Analytical Question) }
\centering
\scalebox{0.80}{
\begin{tabular}{l p{13.5cm}}
&  \includegraphics[width=11.5cm]{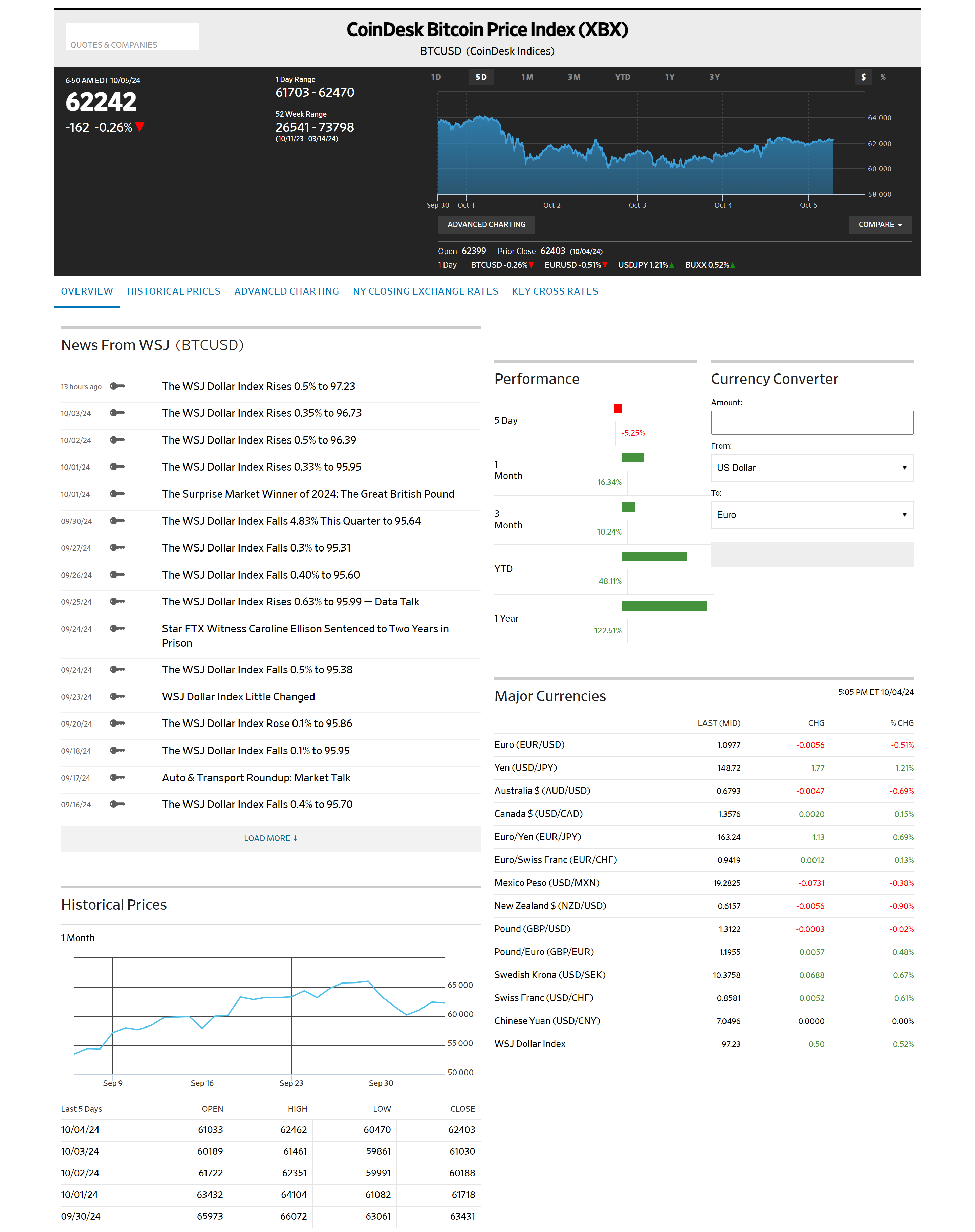} \\
\footnotesize
Question & Based on the provided Bitcoin historical price data, calculate the average closing price for the last five days displayed. \\
\midrule
\footnotesize
LLaVA-NeXT-OV-72B-Chat & To calculate the average closing price for the last five days displayed, we add up the closing prices for each day and then divide by the number of days. The closing prices for the last five days are: 6243, 6103, 6018, 6178, and 62242. Adding these together gives us 6243 + 6103 + 6018 + 6178 + 62242 = 77784. Dividing this sum by 5 gives us an average closing price of 15556.8 for the last five days displayed. \\
GPT4o-Score & 0 \\
GPT4o-Rationale & The assistant's response contains significant errors. It incorrectly lists the closing prices, including an erroneous value of 62242, which is not a closing price but the current price. The correct closing prices are 62403, 60470, 60188, 61718, and 63431. The assistant's summation and division are also incorrect, leading to a final average that is far from the correct value of \$61,642. \\
GPT4o-Response & To calculate the average closing price for the last five days displayed in the provided Bitcoin historical price data, we need to sum the closing prices of those five days and then divide by five. Here are the closing prices for the last five days displayed: 1. 10/04/24: 62403 2. 10/03/24: 61030 3. 10/02/24: 60188 4. 10/01/24: 61718 5. 09/30/24: 63431 Now, let's calculate the average: \[ \text{Average Closing Price} = \frac{62403 + 61030 + 60188 + 61718 + 63431}{5} \] \[ \text{Average Closing Price} = \frac{308770}{5} = 61754 \] So, the average closing price for the last five days displayed is 61,754. \\
\end{tabular}
}
\end{AIbox}
\vspace{-1em}
\captionof{table}{An example of the failure case of LLaVA-NeXT-OV-72B-Chat in LiveBench-09 for Analytical Question}
\vspace{-6mm}
\label{tab:case-LLaVA-OV-Ana}
\end{minipage}
\end{table*}

\begin{table*}[htp]
\begin{minipage}{0.99\textwidth}
\begin{AIbox}{Question Type: Concrete Recognition}
\centering
\scalebox{0.80}{
\begin{tabular}{l p{13.5cm}}
&  \includegraphics[width=13.5cm]{figures/livebench_09/LiveBench_09_sports_tennis.png} \\
\footnotesize
Question & Analyze the ongoing tennis matches displayed on the webpage, detailing the players involved, their current scores, and the tournaments they are part of. \\
\midrule
\footnotesize
Ground Truth & Here are the ongoing tennis matches, scores, and tournaments: 1. China Open: - Karolina Muchova vs. Qinwen Zheng - Current score: 1-1 in the first set. 2. Hong Kong 125 Open: - Anna Blinkova vs. Ajla Tomljanovic - Current score: 7-1 in favor of Anna Blinkova in the first set. 3. Rolex Shanghai Masters: - Zizou Bergs vs. Grigor Dimitrov - Current score: 3-4, match delayed. 4. Rolex Shanghai Masters: - Taylor Fritz vs. Terence Atmane - Current score: 7-4, match delayed. 5. Rolex Shanghai Masters: - Roman Safiullin vs. Alexander Bublik - Current score: 2-1, match delayed. Upcoming matches at 6:30 PM GMT+8: - Nathaniel Lammons/Jackson Withrow vs. Francisco Cerundolo/Tomas Martin Etcheverry - Marcos Giron vs. Karen Khachanov \\
\end{tabular}
}
\end{AIbox}
\captionof{table}{An example of Concrete Recognition question in LiveBench-09}
\vspace{-6mm}
\label{tab:question_understanding}
\end{minipage}
\end{table*}

\begin{table*}[htp]
\begin{minipage}{0.99\textwidth}
\begin{AIbox}{Question Type: Real-World Application}
\centering
\scalebox{0.80}{
\begin{tabular}{l p{13.5cm}}
&  \includegraphics[width=13.5cm]{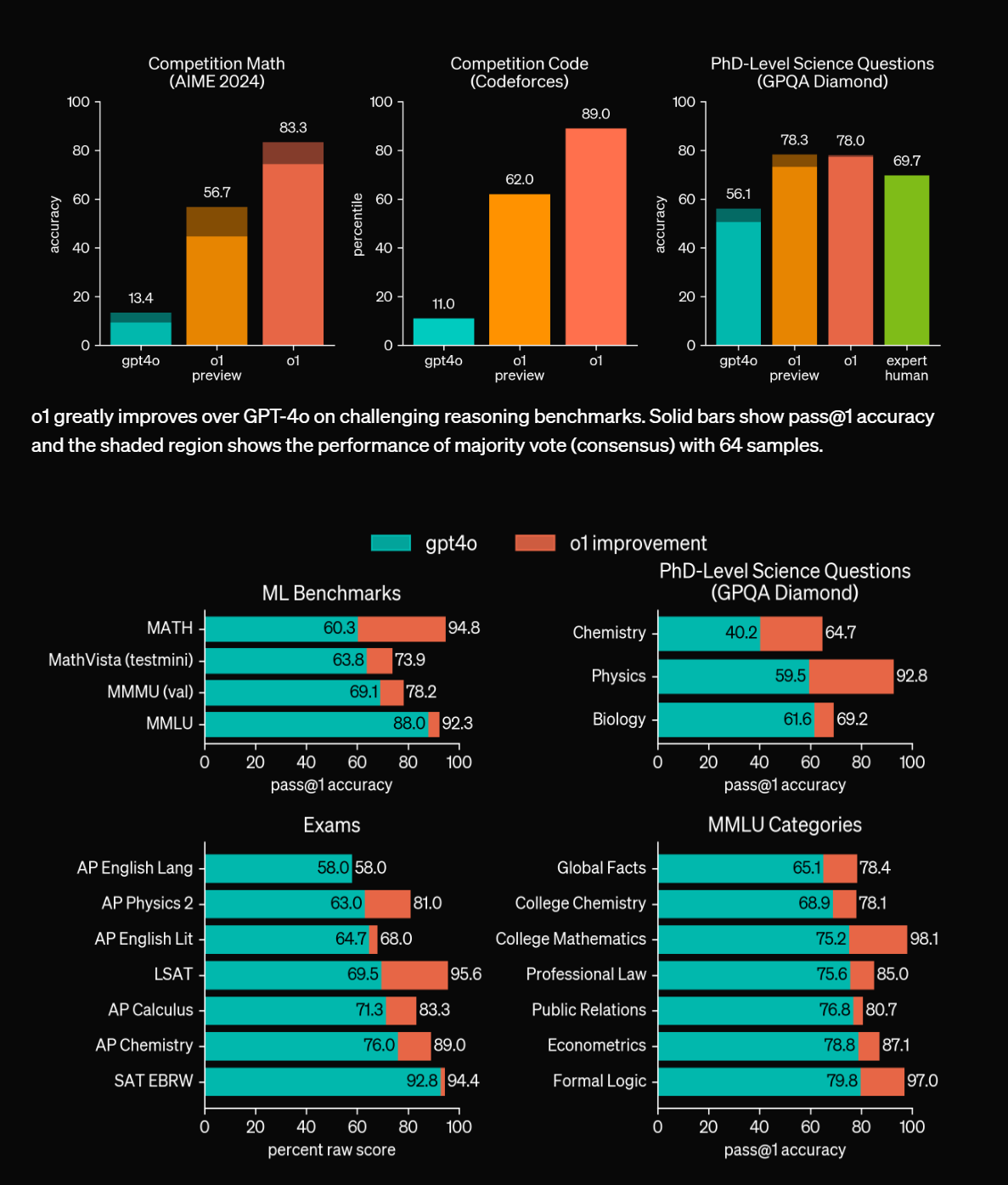} \\
\footnotesize
Question & Create an HTML table summarizing the improvements in 'ML Benchmarks' shown in the image. The table should include the benchmark names, gpt4o scores, and o1 scores. Ensure proper HTML structure and formatting, and include basic styling for better readability. \\
\midrule
\footnotesize
Ground Truth & <!DOCTYPE html> <html lang="en"> <head> <meta charset="UTF-8"> <meta name="viewport" content="width=device-width, initial-scale=1.0"> <title>ML Benchmarks Improvements</title> <style> table { width: 50\%; border-collapse: collapse; margin: 25px 0; font-size: 18px; text-align: left; } th, td { padding: 12px; border-bottom: 1px solid \#ddd; } th { background-color: \#f2f2f2; } </style> </head> <body> <h2>ML Benchmarks Improvements</h2> <table> <thead> <tr> <th>Benchmark</th> <th>gpt4o Score</th> <th>o1 Score</th> </tr> </thead> <tbody> <tr> <td>MATH</td> <td>60.3</td> <td>94.8</td> </tr> <tr> <td>MathVista (testmini)</td> <td>63.8</td> <td>73.9</td> </tr> <tr> <td>MMMU (val)</td> <td>69.1</td> <td>78.2</td> </tr> <tr> <td>MMLU</td> <td>88.0</td> <td>92.3</td> </tr> </tbody> </table> </body> </html> \\
\end{tabular}
}
\end{AIbox}
\captionof{table}{An example of Real World Application question in \texttt{LiveBench-09}}
\vspace{-6mm}
\label{tab:question_application}
\end{minipage}
\end{table*}

\begin{table*}[htp]
\begin{minipage}{0.99\textwidth}
\begin{AIbox}{Question Type: Analytical Understanding}
\centering
\scalebox{0.80}{
\begin{tabular}{l p{13.5cm}}
&  \includegraphics[width=13.5cm]{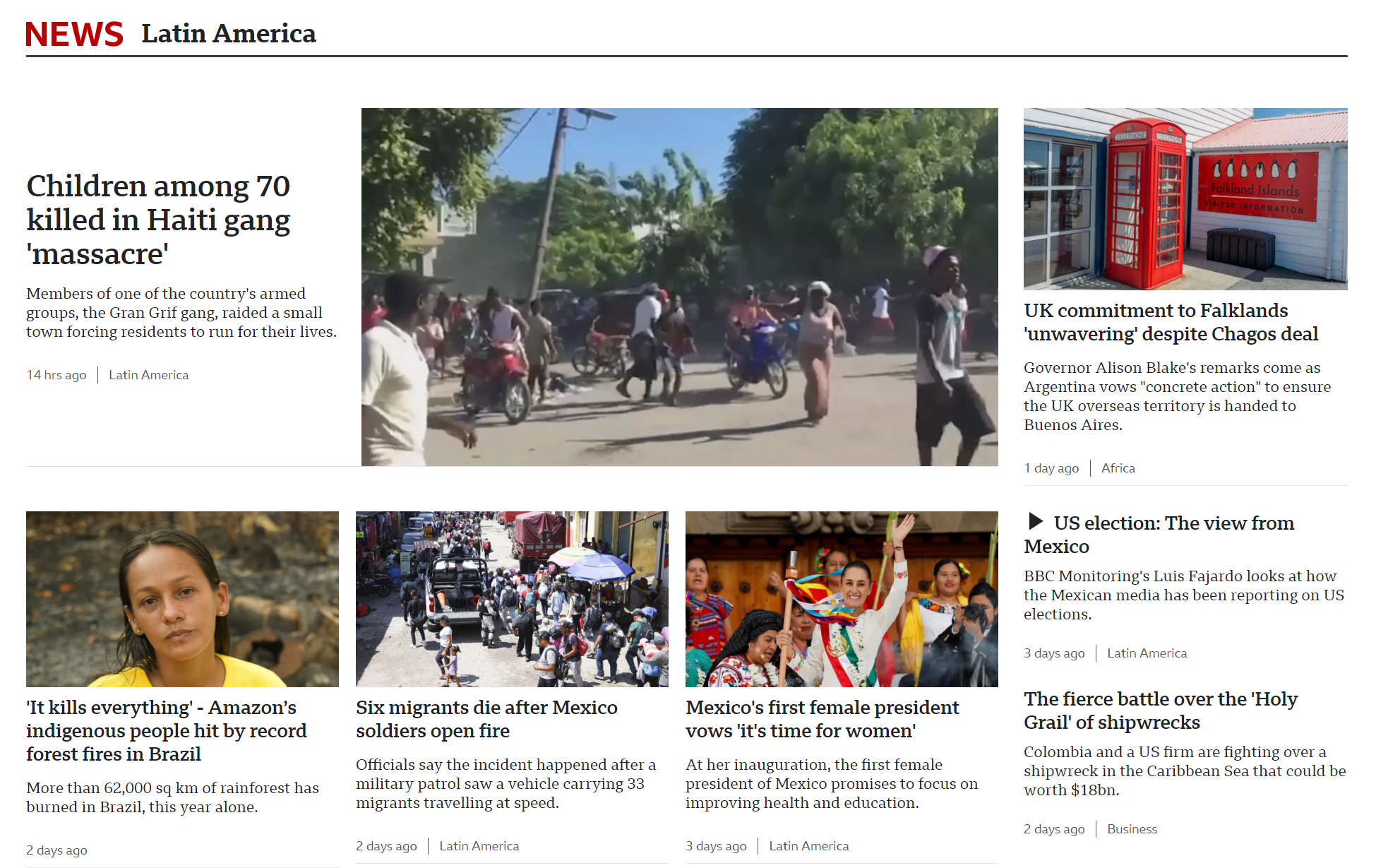} \\
\footnotesize
Question & 	
Analyze the scene depicted in the image associated with the Haiti gang 'massacre' article. Describe the environment, the actions of the people, and the emotions conveyed. How do these elements support the narrative described in the article? \\
\midrule
\footnotesize
Ground Truth & The image related to the Haiti gang 'massacre' article depicts a chaotic environment with frantic activity. People are seen running and some are on motorcycles, suggesting a hurried evacuation. The background shows buildings, trees, and power lines in what seems to be a small town setting. The overall mood is one of panic and urgency, as the townspeople appear distressed and desperate to escape, which aligns with the article's description of residents being forced to flee due to a gang raid by the Gran Grif gang. The image effectively conveys the fear and chaos precipitated by such violent incidents, underscoring the severe impact on the community. \\
\end{tabular}
}
\end{AIbox}
\captionof{table}{An example of Analytical Question in \texttt{LiveBench-09}}
\vspace{-6mm}
\label{tab:question_analytical}
\end{minipage}
\end{table*}

\begin{table*}[htp]
\begin{minipage}{0.99\textwidth}
\begin{AIbox}{Question Type: Divergent Thinking \& Creation}
\centering
\scalebox{0.80}{
\begin{tabular}{l p{13.5cm}}
&  \includegraphics[width=13.5cm]{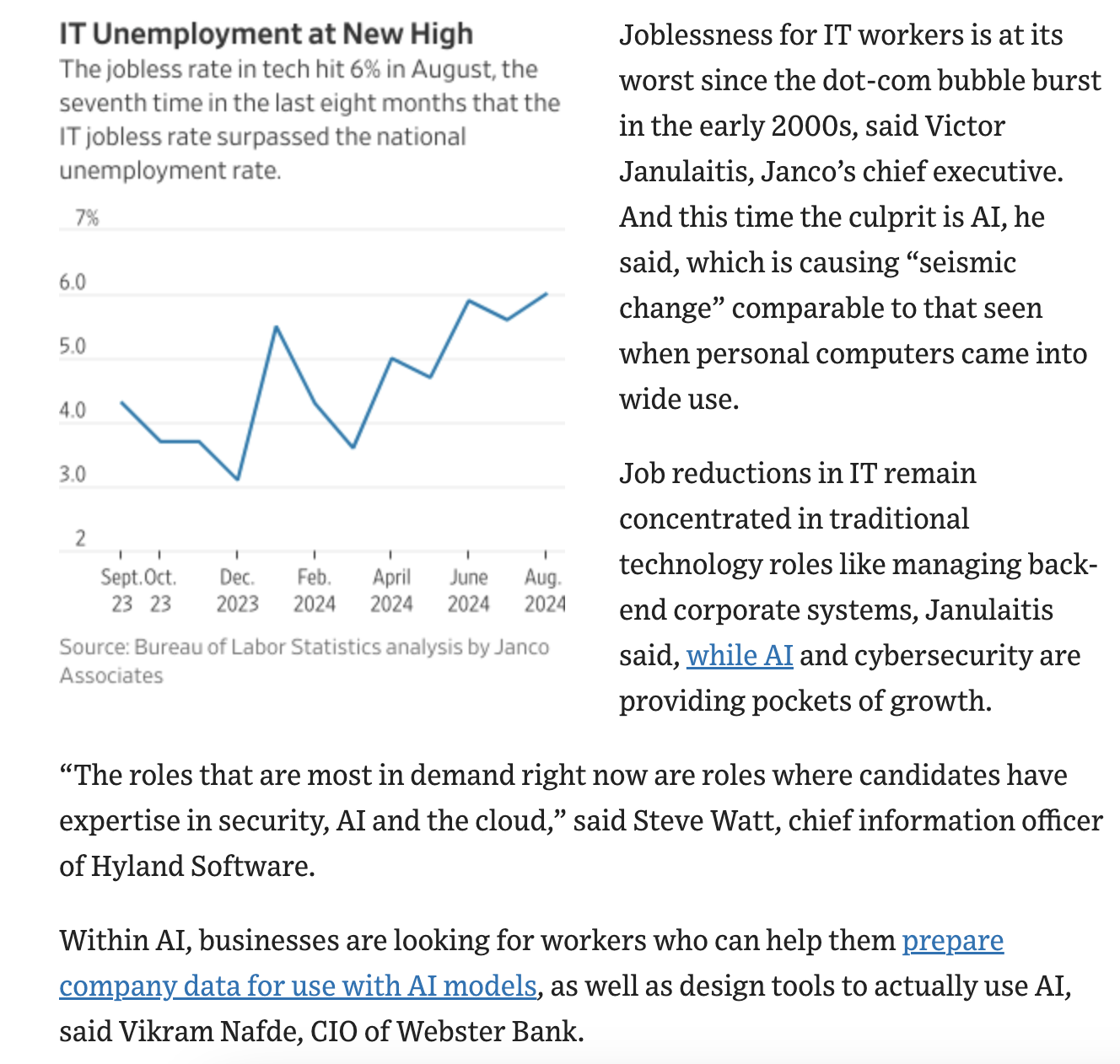} \\
\footnotesize
Question & Evaluate the potential impact of AI on the IT job market, considering both the negative and positive effects described in the provided image and text. Discuss the short-term and long-term impacts, and suggest strategies for IT professionals to adapt to these changes. \\
\midrule
\footnotesize
Ground Truth & The potential impact of AI on the IT job market is multifaceted, with both negative and positive effects, as illustrated by the provided image and text. **Negative Effects:** 1. **Job Reductions in Traditional Roles:** - The IT jobless rate rose to 6\% in August, surpassing the national unemployment rate multiple times over the past eight months. This suggests a significant disruption in traditional IT jobs, such as managing back-end corporate systems, due to AI's capabilities in automating and optimizing these tasks. 2. **Seismic Changes:** - Victor Janulaitis, Janco's chief executive, compares the impact of AI to the seismic changes seen when personal computers came into wide use. This kind of systemic shift often results in skills becoming obsolete, leading to higher unemployment among professionals who cannot quickly adapt to new technological paradigms. **Positive Effects:** 1. **Growth in AI and Cybersecurity Roles:** - Despite the job reductions in some areas, AI and cybersecurity are cited as providing pockets of growth. This indicates that as AI takes over routine tasks, there is increased demand for developing, managing, and securing AI systems. 2. **High Demand for Expertise in Security, AI, and the Cloud:** - Steve Watt of Hyland Software mentions that roles requiring expertise in security, AI, and the cloud are most in-demand. This reflects a shift in the skill sets that are valuable in the job market towards more advanced and specialized IT roles. 3. **New Roles in AI Model Preparation and Tool Design:** - Vikram Nafde of Webster Bank highlights the need for workers to prepare company data for AI models and design tools to leverage AI. These new and emerging roles provide opportunities for IT professionals to transition into more specialized fields that AI technology creates.  ... \{Remaining Ground Truth answer\}\\
\end{tabular}
}
\end{AIbox}
\captionof{table}{An example of Creation Question in \texttt{LiveBench-09}}
\vspace{-6mm}
\label{tab:question_creation}
\end{minipage}
\end{table*}

\begin{table*}[h]
\caption{Dataset Statistics in \lmmseval. This table categorizes the initial set of tasks, detailing their task domains, ground-truth types, instance counts, and splits. We provide a comprehensive overview of the diverse datasets employed, which cover various task domains and evaluation metrics.}
\label{tab:lmms_eval_dataset}
\centering
\resizebox{\textwidth}{!}{%
\begin{tabular}{@{}l|l|l|l|l@{}}
\toprule
\textbf{Datasets} & \textbf{Task Domains} & \textbf{Ground-Truth Types} & \textbf{Instances} & \textbf{Splits} \\ \midrule
AI2D \citep{kembhavi2016diagram} & Science,Diagram & Muiti-Choice & 3088 & test \\
BenchLMM ~\citep{cai2023benchlmm} & Cross Style Understanding & Short Answer / Muiti-Choice & 102 & test \\
ChartQA \citep{masry2022chartqa} & Chart & Short Answer & 2500 & test \\
CMMMU \citep{zhang2024cmmmu} & Multi-task,World Knowledge & Free-form / Muiti-Choice & 900/11000 & val/test \\
COCO 2014 Caption \citep{lin2014microsoft} & Captioning & Short Answer & 40775 / 40504 & test / val \\
COCO 2017 Caption \citep{lin2014microsoft} & Captioning & Short Answer & 40670 / 5000 & test / val \\
DocVQA \citep{mathew2020docvqa} & Document & Short Answer & 5349 & test \\
Ferret \citep{you2023ferret} & Referring or Grounding Actions & Free-form Answer & 120 & test \\
Flickr30k \citep{young2014image} & Visual Understanding & Captioning & 31783 & test \\
GQA \citep{hudson2019gqa} & Real-World/Compositional QA & Short Answer & 12578 & test / dev \\
Hallusion-Bench \citep{guan2023hallusionbench} & Multimodal Image-Context Reasoning & Yes or No & 951 & image \\
IconQA \citep{lu2022iconqa} & Abstract Diagrams & Muiti-Choice / Short Answer & 21489 / 21488 & test / val \\
InfoVQA \citep{Mathew_2022_WACV} & Infographics understanding & Extractive / Numerical & 2801 & val \\
LLaVA-COCO \citep{liu2023visual} & Conversation, Reasoning & Free-form Answer & 90 & test \\
LLaVA-W \citep{liu2023visual} & Conversation, Reasoning & Free-form Answer & 60 & test \\
LLaVA-Wilder \citep{liu2024llava} & Conversation, Reasoning & Free-form Answer & 210/1020 & test \\
LiveBench (Ours) & Webpage Understanding / Lively Updated & Free-form & dynamic & test \\
MathVista \citep{lu2024mathvista} & Mathematical Reasoning / Understanding & Free-form / Muiti-Choice & 1000 & testmini \\
MathVerse \citep{zhang2024mathverse} & Mathematical Reasoning / Understanding & Free-form / Muiti-Choice & 3940 & testmini \\
MMBench \citep{liu2023mmbench} & Reasoning / Perception & Muiti-Choice & 6666 / 4329 & test / dev \\
MME \citep{fu2024mme} & Perception, Cognition & Yes or No & 2374 & test \\
MMMU \citep{yue2023mmmu} & Multi-task, World Knowledge & Free-form / Muiti-Choice & 10500 / 900 & test / val \\
MM-Vet \citep{yu2023mmvet} & Multi-task & Free-form & 218 & test \\
Multilingual-LLaVA-W & Multi-lingual Conversation,Reasoning & Free-form Answer & 60 & test \\
MultiDocVQA \citep{tito2023hierarchical} & Document & Short Answer & 5019 / 5187 & test / val \\
NoCaps \citep{agrawal2019nocaps} & Novel Object Captioning & Short Answer & 4500 & val \\
OCRBench \citep{liu2023hidden} & Text Recognition & Short Answer & 1000 & test \\
OKVQA \citep{marino2019ok} & knowledge-based visual QA & Short Answer & 5046 & val \\
OlympiadBench \citep{he2024olympiadbench} & Reasoning & Short Answer & 2126 / 6351 & test-en / test-cn \\
POPE \citep{li2023evaluating} & Hallucination & Yes or No & 9000 & test \\
Q-Bench ~\citep{wu2023q} & Image Quality Assessment & Short Answer / Muiti-Choice & 2990 & test \\
RealWorldQA ~\citep{grokv2024} & Real world scenarios QA & Muiti-Choice & 765 & test \\
Refcoco \citep{kazemzadeh-etal-2014-referitgame, Mao_2016_CVPR} & Referring Expression & Short Answer & 5000 / 1975 / 1810 / 8811 & bbox-test / A / B / val \\
Refcoco \citep{kazemzadeh-etal-2014-referitgame, Mao_2016_CVPR} & Referring Expression & Short Answer & 5000 / 1975 / 1810 / 8811 & seg-test / A / B / val \\
Refcoco+ \citep{kazemzadeh-etal-2014-referitgame, Mao_2016_CVPR} & Referring Expression & Short Answer & 1975 / 1798 / 3805, & bbox-testA / B / val \\
Refcoco+ \citep{kazemzadeh-etal-2014-referitgame, Mao_2016_CVPR} & Referring Expression & Short Answer & 1975 / 1798 / 3805 & seg-testA / B / val \\
Refcocog \citep{kazemzadeh-etal-2014-referitgame, Mao_2016_CVPR} & Referring Expression & Short Answer & 5023 / 7573 & bbox-testB / val, \\
Refcocog \citep{kazemzadeh-etal-2014-referitgame, Mao_2016_CVPR} & Referring Expression & Short Answer & 5023 / 7573 & seg-test / val \\
ScienceQA \citep{lu2022learn} & Science, World Knowledge, Reasoning & Muiti-Choice & 4241 & test \\
ScreenSPOT~\citep{cheng2024seeclick} & GUI Understanding / Navigation & Short Answer / Coordinates & 1272 & test \\
SEED-Bench \citep{li2023seed} & Spatial and Temporal Understanding & Muiti-Choice & 17990 & test \\
SEED-Bench-2 \citep{li2023seedbench2} & Multi-disciplinary Knowledge & Muiti-Choice & 24371 & test \\
ST-VQA \citep{biten2019scene} & Highlevel Semantic Information Understanding & Short Answer & 4070 & test \\
SynthDoG \citep{kim2022donut} & Text Understanding & Free-form & 500 / 500 & val-en / val-zh \\
TextCaps \citep{sidorov2019textcaps} & Text Understanding & Captioning & 21953 / 3166 / 3289 & train / val / test \\
TextVQA \citep{singh2019towards} & Text Understanding & Short Answer & 5000 / 5734 & val / test \\
VisualWebBench~\citep{liu2024visualwebbench} & Webpage Understanding / OCR / Reasoning & Short Answer / Muiti-Choice & 1536 & test \\
VizwizVQA \citep{gurari2018vizwiz} & Low Quality Image Understanding & Short Answer & 8000 / 4319 & test / val \\
VQAv2 \citep{balanced_vqa_v2} & Visual QA & Free-form & 447793 / 214354 & test / val \\
WebSRC \citep{chen2021websrc} & Structure of Webpage & Short Answer / Yes or No & 40357 / 52826 & test / dev \\ \bottomrule
\end{tabular}%
}
\end{table*}

\section{\lmmseval~Suite Information}
\label{appendix:suite-info}

\begin{table*}[h]
    \centering
    \caption{Detailed Statistics of the Initial Set of Models in \lmmseval. The models are categorized by their model family, with their inference parameters, model types (indicating whether they are open-sourced or accessed via API), and parallel types, which denote the strategy leveraged during the model inference.}
    \label{tab:lmms_eval_models}
    \setlength{\tabcolsep}{8pt}
\renewcommand{\arraystretch}{1.2}
\resizebox{0.80\textwidth}{!}{%
\begin{tabular}{@{}ccccc@{}}
\toprule
\textbf{Model Family} & \textbf{Model Version} & \textbf{Parameters} & \textbf{Model Type} & \textbf{Parallel Type} \\ \midrule
\multicolumn{1}{c|}{\multirow{2}{*}{InstructBLIP}} & \multicolumn{1}{c|}{InstructBLIP-Vicuna-7B} & \multicolumn{1}{c|}{7B} & \multicolumn{1}{c|}{Open-sourced} & Data \\
\multicolumn{1}{c|}{} & \multicolumn{1}{c|}{InstructBLIP-Vicuna-13B} & \multicolumn{1}{c|}{13B} & \multicolumn{1}{c|}{Open-sourced} & Data \\ \midrule
\multicolumn{1}{c|}{Fuyu} & \multicolumn{1}{c|}{Fuyu-8B} & \multicolumn{1}{c|}{8B} & \multicolumn{1}{c|}{Open-sourced} & Data \\ 
\multicolumn{1}{c|}{Idefics} & \multicolumn{1}{c|}{Idefics-2-8B} & \multicolumn{1}{c|}{8B} & \multicolumn{1}{c|}{Open-sourced} & Data \\ 
\multicolumn{1}{c|}{MiniCPM} & \multicolumn{1}{c|}{MiniCPM-V 2.8B} & \multicolumn{1}{c|}{2.8B} & \multicolumn{1}{c|}{Open-sourced} & Data \\ 
\multicolumn{1}{c|}{XComposer} & \multicolumn{1}{c|}{XComposer-4KHD} & \multicolumn{1}{c|}{8B} & \multicolumn{1}{c|}{Open-sourced} & Data \\ 
\multicolumn{1}{c|}{InternVL} & \multicolumn{1}{c|}{InternVL-1.5} & \multicolumn{1}{c|}{26B} & \multicolumn{1}{c|}{Open-sourced} & Data \\ \midrule
\multicolumn{1}{c|}{\multirow{9}{*}{LLaVA}} & \multicolumn{1}{c|}{LLaVA-1.5-7B} & \multicolumn{1}{c|}{7B} & \multicolumn{1}{c|}{Open-sourced} & Data \\
\multicolumn{1}{c|}{} & \multicolumn{1}{c|}{LLaVA-1.5-13B} & \multicolumn{1}{c|}{13B} & \multicolumn{1}{c|}{Open-sourced} & Data \\
\multicolumn{1}{c|}{} & \multicolumn{1}{c|}{LLaVA-NeXT-Vicuna-7B} & \multicolumn{1}{c|}{7B} & \multicolumn{1}{c|}{Open-sourced} & Data \\
\multicolumn{1}{c|}{} & \multicolumn{1}{c|}{LLaVA-NeXT-Vicuna-13B} & \multicolumn{1}{c|}{13B} & \multicolumn{1}{c|}{Open-sourced} & Data \\
\multicolumn{1}{c|}{} & \multicolumn{1}{c|}{LLaVA-NeXT-Mistral-7B} & \multicolumn{1}{c|}{7B} & \multicolumn{1}{c|}{Open-sourced} & Data \\
\multicolumn{1}{c|}{} & \multicolumn{1}{c|}{LLaVA-NeXT-Yi-34B} & \multicolumn{1}{c|}{34B} & \multicolumn{1}{c|}{Open-sourced} & Data \\
\multicolumn{1}{c|}{} & \multicolumn{1}{c|}{LLaVA-NeXT-LLaMA-3-8B} & \multicolumn{1}{c|}{8B} & \multicolumn{1}{c|}{Open-sourced} & Data \\
\multicolumn{1}{c|}{} & \multicolumn{1}{c|}{LLaVA-NeXT-Qwen-72B} & \multicolumn{1}{c|}{72B} & \multicolumn{1}{c|}{Open-sourced} & Model \\
\multicolumn{1}{c|}{} & \multicolumn{1}{c|}{LLaVA-NeXT-Qwen-110B} & \multicolumn{1}{c|}{110B} & \multicolumn{1}{c|}{Open-sourced} & Model \\ \midrule
\multicolumn{1}{c|}{\multirow{3}{*}{Qwen-VL}} & \multicolumn{1}{c|}{Qwen-VL-Chat-7B} & \multicolumn{1}{c|}{7B} & \multicolumn{1}{c|}{Open-sourced} & Data \\
\multicolumn{1}{c|}{} & \multicolumn{1}{c|}{Qwen-VL-Plus} & \multicolumn{1}{c|}{N/A} & \multicolumn{1}{c|}{Close-sourced, API} & Data \\
\multicolumn{1}{c|}{} & \multicolumn{1}{c|}{Qwen-VL-MAX} & \multicolumn{1}{c|}{N/A} & \multicolumn{1}{c|}{Close-sourced, API} & Data \\ \midrule
\multicolumn{1}{c|}{\multirow{3}{*}{Gemini}} & \multicolumn{1}{c|}{Gemini-1.0-Pro} & \multicolumn{1}{c|}{N/A} & \multicolumn{1}{c|}{Close-sourced, API} & Data \\
\multicolumn{1}{c|}{} & \multicolumn{1}{c|}{Gemini-1.5-Flash} & \multicolumn{1}{c|}{N/A} & \multicolumn{1}{c|}{Close-sourced, API} & Data \\
\multicolumn{1}{c|}{} & \multicolumn{1}{c|}{Gemini-1.5-Pro} & \multicolumn{1}{c|}{N/A} & \multicolumn{1}{c|}{Close-sourced, API} & Data \\ \midrule
\multicolumn{1}{c|}{\multirow{2}{*}{GPT4}} & \multicolumn{1}{c|}{GPT-4V} & \multicolumn{1}{c|}{N/A} & \multicolumn{1}{c|}{Close-sourced, API} & Data \\
\multicolumn{1}{c|}{} & \multicolumn{1}{c|}{GPT-4O} & \multicolumn{1}{c|}{N/A} & \multicolumn{1}{c|}{Close-sourced, API} & Data \\ \midrule
\multicolumn{1}{c|}{\multirow{3}{*}{Claude}} & \multicolumn{1}{c|}{Claude-3-Haku} & \multicolumn{1}{c|}{N/A} & \multicolumn{1}{c|}{Close-sourced, API} & Data \\
\multicolumn{1}{c|}{} & \multicolumn{1}{c|}{Claude-3-Sonnet} & \multicolumn{1}{c|}{N/A} & \multicolumn{1}{c|}{Close-sourced, API} & Data \\
\multicolumn{1}{c|}{} & \multicolumn{1}{c|}{Claude-3-Opus} & \multicolumn{1}{c|}{N/A} & \multicolumn{1}{c|}{Close-sourced, API} & Data \\ \bottomrule
\end{tabular}%
}
\end{table*}

\textbf{Datasets on LMMs-Eval} In previous research, benchmarks such as AI2D~\citep{kembhavi2016diagram}, TextVQA~\citep{singh2019vqa}, TextCaps~\citep{sidorov2020textcaps}, Flickr30k~\citep{young-etal-2014-image}, and OK-VQA~\citep{okvqa} among many others, have been employed to assess a model’s performance in tasks such as captioning, optical character recognition (OCR), and visual QA. With the advent of Large Multimodal Models (LMMs), these have increasingly focused on broader capabilities spanning both vision and language, including reasoning~\citep{lu2022learn} and visual instruction following~\citep{liu2023visual}. Consequently, new benchmarks featuring increasingly challenging tasks and more comprehensive evaluations have been proposed. For example, ScienceQA~\citep{lu2022learn} and MathVista~\citep{lu2024mathvista} assess mathematical and scientific competencies, while benchmarks like SEED-Bench~\citep{li2023seedbench}, CMMMU~\citep{zhang2024cmmmu}, MMMU~\citep{yue2023mmmu}, and MM-Bench~\citep{liu2024mmbench} evaluate the multifaceted dimensions of multimodal models.


\textbf{Models on LMMs-Eval} To enable comparisons on new benchmarks for different models and to understand their capabilities across multiple tasks, we have supported over 10 models such as Fuyu~\citep{fuyu-8b}, LLaVA~\citep{liu2023visual}, Instruct-BLIP~\citep{dai2023instructblip}, InternVL~\citep{chen2023internvl}, XComposer~\citep{dong2024internlm}, Qwen-VL~\citep{bai2023qwenvl}, MiniCPM~\citep{viscpm}, Idefics~\citep{laurençon2024matters} and closed-source models such as GPT-4V~\citep{2023GPT4VisionSC}, Gemini~\citep{geminiteam2024gemini}, Qwen-VL-Max~\citep{qwenvl2024} and Claude~\citep{claude3family2024}.

\subsection{Unified Evaluation Results with \lmmseval}
\label{appendix:lmms-eval-more}

\begin{table*}[h]
    \centering
    \caption{More results using \lmmseval~}
    \label{tab:more-lmms-eval-results}
    \resizebox{\textwidth}{!}{

\begin{tabular}{@{}l|ccc|cccccc@{}}

\toprule

            & Split & Metric   & \#Num & LLaVA-1.5-7B & LLaVA-1.5-13B & LLaVA-NeXT-mistral-7B & LLaVA-NeXT-vicuna-7B & LLaVA-NeXT-13B & LLaVA-NeXT-34B \\

\midrule

COCO-Cap     & cococap\_val\_2014              & CIDEr        & 40,504  & 108.66 & 113.88 & 107.66 & 96.98 & 99.45  & 103.16 \\
COCO-Cap     & cococap\_val\_2017              & CIDEr        & 5,000   & 110.38 & 115.61 & 109.22 & 99.93 & 101.99 & 105.89 \\
DocVQA       & val                             & ANLS         & 5,349   & 28.08  & 30.29  & 72.16  & 74.35 & 77.45  & 83.98  \\
GQA          & testdev\_balanced\_instructions & Acc          & 12,578  & 61.97  & 63.24  & 54.98  & 64.23 & 65.36  & 67.08  \\
MultidocVQA & val   & Anls/acc & 5,187 & 16.65/7.21   & 18.25/8.02    & 41.4/27.89            & 44.42/31.32          & 46.28/32.56    & 50.16/34.93    \\
NoCaps       & nocaps\_eval                    & CIDEr        & 4,500   & 105.54 & 109.28 & 96.14  & 88.29 & 88.27  & 91.94  \\
OKVQA        & val                             & Acc          & 5,046   & 53.44  & 58.22  & 54.77  & 44.25 & 46.27  & 46.84  \\
POPE         & test                            & F1 Score     & 9,000   & 85.87  & 85.92  & 86.79  & 86.4  & 86.26  & 87.77  \\
ScienceQA    & scienceqa-full                  & Acc.         & 4,114   & 70.41  & 74.96  & 28.84  & 73.21 & 75.85  & 85.81  \\
Refcoco      & all                             & CIder        & 17,596  & 29.76  & 34.26  & 9.47   & 34.2  & 34.75  & 33.56  \\
Refcoco+     & all                             & CIder        & 7,578   & 28.92  & 31.01  & 9.05   & 31.82 & 32     & 30.66  \\
Refcocog     & all                             & CIder        & 12,596  & 57.76  & 59.23  & 19.35  & 52.18 & 58.02  & 59.26  \\
ScienceQA    & scienceqa-img                   & Acc          & 2,017   & 70.43  & 72.88  & 28.56  & 70.15 & 73.57  & 81.85  \\
SEED-Bench   & Seed-1                          & Image-Acc    & 17,990  & 60.49  & 67.06  & 65.97  & 64.74 & 65.64  & 69.55  \\
SEED-Bench-2 & Seed-2                          & Acc          & 24,371  & 57.89  & 59.88  & 60.83  & 59.88 & 60.72  & 64.98  \\
TextCaps     & val                             & CIDEr        & 3,166   & 98.15  & 103.92 & 70.39  & 71.79 & 67.39  & 67.11  \\
TextVQA      & val                             & exact\_match & 5,000   & 46.07  & 48.73  & 65.76  & 64.85 & 66.92  & 69.31  \\
VizWiz(val)  & val                             & Acc          & 4,319   & 54.39  & 56.65  & 63.79  & 60.64 & 63.56  & 66.61  \\
VQAv2        & val                             & Acc          & 214,354 & 76.64  & 78.26  & 80.32  & 80.06 & 80.92  & 82.07 \\

\bottomrule
\end{tabular}

}
\end{table*}

We present additional results using \lmmseval~ here. Due to limited computational resources, we are only able to provide a holistic view of models from the LLaVA \citep{liu2023improvedllava} series. This demonstrates that achieving both wide coverage and low-cost evaluation simultaneously is not feasible, necessitating a balance between these two aspects.

\begin{table*}[h]
    \centering
    \caption{List of websites selected for \livebench. }
    \label{tab:websites}
    \setlength{\tabcolsep}{2pt}
    \renewcommand{\arraystretch}{1.0}
    \resizebox{0.85\textwidth}{!}{
        \begin{tabular}{@{}l|l|l@{}}
            \toprule
            \textbf{Name} & \textbf{URL} & \textbf{Category} \\ \midrule
            BBC Main & \url{https://www.bbc.com/} & General News \\ \hline
            BBC News & \url{https://www.bbc.com/news} & News \\ \hline
            BBC Sport & \url{https://www.bbc.com/sport} & Sports \\ \hline
            BBC Business & \url{https://www.bbc.com/business} & Business \\ \hline
            BBC Innovation & \url{https://www.bbc.com/innovation} & Innovation \\ \hline
            BBC Culture & \url{https://www.bbc.com/culture} & Culture \\ \hline
            BBC Travel & \url{https://www.bbc.com/travel} & Travel \\ \hline
            BBC Future Planet & \url{https://www.bbc.com/future-planet} & Environment \\ \hline
            CNN Main & \url{https://edition.cnn.com/} & General News \\ \hline
            CNN Politics & \url{https://edition.cnn.com/politics} & Politics \\ \hline
            CNN Entertainment & \url{https://edition.cnn.com/entertainment} & Entertainment \\ \hline
            CNN Style & \url{https://edition.cnn.com/style} & Style \\ \hline
            Bloomberg Economics & \url{https://www.bloomberg.com/economics} & Economics \\ \hline
            Bloomberg Industries & \url{https://www.bloomberg.com/industries} & Industries \\ \hline
            Bloomberg Technology & \url{https://www.bloomberg.com/technology} & Technology \\ \hline
            Bloomberg Politics & \url{https://www.bloomberg.com/politics} & Politics \\ \hline
            Bloomberg Opinion & \url{https://www.bloomberg.com/opinion} & Opinion \\ \hline
            WSJ Main & \url{https://www.wsj.com/} & General News \\ \hline
            WSJ Africa & \url{https://www.wsj.com/world/africa?mod=nav_top_subsection} & Africa \\ \hline
            WSJ Americas & \url{https://www.wsj.com/world/americas?mod=nav_top_subsection} & Americas \\ \hline
            WSJ Asia & \url{https://www.wsj.com/world/asia?mod=nav_top_subsection} & Asia \\ \hline
            WSJ China & \url{https://www.wsj.com/world/china?mod=nav_top_subsection} & China \\ \hline
            WSJ Europe & \url{https://www.wsj.com/world/europe?mod=nav_top_subsection} & Europe \\ \hline
            WSJ Middle East & \url{https://www.wsj.com/world/middle-east?mod=nav_top_subsection} & Middle East \\ \hline
            WSJ India & \url{https://www.wsj.com/world/india?mod=nav_top_subsection} & India \\ \hline
            WSJ Oceania & \url{https://www.wsj.com/world/oceania?mod=nav_top_subsection} & Oceania \\ \hline
            WSJ Russia & \url{https://www.wsj.com/world/russia?mod=nav_top_subsection} & Russia \\ \hline
            WSJ UK & \url{https://www.wsj.com/world/uk?mod=nav_top_subsection} & UK \\ \hline
            WSJ Science & \url{https://www.wsj.com/science?mod=nav_top_subsection} & Science \\ \hline
            WSJ Archaeology & \url{https://www.wsj.com/science/archaeology?mod=nav_top_subsection} & Archaeology \\ \hline
            WSJ Biology & \url{https://www.wsj.com/science/biology?mod=nav_top_subsection} & Biology \\ \hline
            WSJ Environment & \url{https://www.wsj.com/science/environment?mod=nav_top_subsection} & Environment \\ \hline
            WSJ Physics & \url{https://www.wsj.com/science/physics?mod=nav_top_subsection} & Physics \\ \hline
            WSJ Space & \url{https://www.wsj.com/science/space-astronomy?mod=nav_top_subsection} & Space \\ \hline
            WSJ Central Banking & \url{https://www.wsj.com/economy/central-banking?mod=nav_top_subsection} & Central Banking \\ \hline
            WSJ Consumers & \url{https://www.wsj.com/economy/consumers?mod=nav_top_subsection} & Consumers \\ \hline
            WSJ Housing & \url{https://www.wsj.com/economy/housing?mod=nav_top_subsection} & Housing \\ \hline
            WSJ Jobs & \url{https://www.wsj.com/economy/jobs?mod=nav_top_subsection} & Jobs \\ \hline
            WSJ Trade & \url{https://www.wsj.com/economy/trade?mod=nav_top_subsection} & Trade \\ \hline
            WSJ Global & \url{https://www.wsj.com/economy/global} & Global Economy \\ \hline
            WSJ AI & \url{https://www.wsj.com/tech/ai?mod=nav_top_subsection} & AI \\ \hline
            WSJ Biotech & \url{https://www.wsj.com/tech/biotech} & Biotech \\ \hline
            WSJ Cybersecurity & \url{https://www.wsj.com/tech/cybersecurity?mod=nav_top_subsection} & Cybersecurity \\ \hline
            WSJ Personal Tech & \url{https://www.wsj.com/tech/personal-tech?mod=nav_top_subsection} & Personal Tech \\ \hline
            Reuters Main & \url{https://www.reuters.com/} & General News \\ \hline
            Reuters Aerospace and Defense & \url{https://www.reuters.com/business/aerospace-defense/} & Aerospace and Defense \\ \hline
            Reuters Autos and Transportation & \url{https://www.reuters.com/business/autos-transportation/} & Autos and Transportation \\ \hline
            Reuters Davos & \url{https://www.reuters.com/business/davos/} & Davos \\ \hline
            Reuters Energy & \url{https://www.reuters.com/business/energy/} & Energy \\ \hline
            Reuters Environment & \url{https://www.reuters.com/business/environment/} & Environment \\ \hline
            Reuters Finance & \url{https://www.reuters.com/business/finance/} & Finance \\ \hline
            Reuters Healthcare & \url{https://www.reuters.com/business/healthcare-pharmaceuticals/} & Healthcare\\ \hline
            Reuters Media and Telecom & \url{https://www.reuters.com/business/media-telecom/} & Media and Telecom \\ \hline
            Reuters Retail and Consumer & \url{https://www.reuters.com/business/retail-consumer/} & Retail and Consumer \\ \hline
            Reuters Future of Health & \url{https://www.reuters.com/business/future-of-health/} & Future of Health \\ \hline
            Reuters Future of Money & \url{https://www.reuters.com/business/future-of-money/} & Future of Money \\ \hline
            Reuters Take Five & \url{https://www.reuters.com/business/take-five/} & Analysis \\ \hline
            Reuters World at Work & \url{https://www.reuters.com/business/world-at-work/} & World at Work \\ \hline
            Reuters Breakingviews & \url{https://www.reuters.com/breakingviews/} & Opinion \\ \hline
            Reuters Technology & \url{https://www.reuters.com/technology/} & Technology \\ \hline
            Reuters Cybersecurity & \url{https://www.reuters.com/technology/cybersecurity/} & Cybersecurity \\ \hline
            Reuters Space & \url{https://www.reuters.com/technology/space/} & Space \\ \hline
            Reuters Disrupted & \url{https://www.reuters.com/technology/disrupted/} & Disruption \\ \hline
            Reuters Momentum & \url{https://www.reuters.com/technology/reuters-momentum/} & Technology \\ \hline
            Reuters Investigations & \url{https://www.reuters.com/investigations/} & Investigations \\ \hline
            Andreessen Horowitz & \url{https://a16z.com/news-content/\#latest} & Technology \\ \hline
            Hacker News & \url{https://news.ycombinator.com/} & Technology \\ \hline
            Reddit & \url{https://www.reddit.com/?rdt=48006} & Social Media \\ \hline
            Crunchbase News & \url{https://news.crunchbase.com/} & Startups \\ \hline
            CCTV & \url{https://www.cctv.com/} & International News \\ \bottomrule
        \end{tabular}
    }
\end{table*}

\end{document}